\documentclass[journal]{IEEEtran}
\usepackage{amsmath, amssymb, amsthm}
\usepackage{algorithm}
\usepackage{algpseudocode}
\usepackage{algorithm}
\usepackage{algpseudocode}
\usepackage{graphicx}
\usepackage{subcaption}
\usepackage{setspace}
\usepackage{multicol}
\usepackage[font=small,labelfont=bf]{caption}
\usepackage{listings}
\usepackage{verbatim}
\usepackage{fancyvrb}
\usepackage{adjustbox}
\usepackage{xcolor}
\usepackage{tabularx}  
\usepackage[normalem]{ulem}
\usepackage[utf8]{inputenc}
\usepackage{url}
\usepackage[
    colorlinks=true,
    linkcolor=blue,
    citecolor=blue,
    urlcolor=blue
]{hyperref}
\usepackage{cleveref}
\usepackage{xcolor}
\def\BibTeX{{\rm B\kern-.05em{\sc i\kern-.025em b}\kern-.08em
    T\kern-.1667em\lower.7ex\hbox{E}\kern-.125emX}}

\begin{document}

\title{Securing Time Integrity in Energy IoT Against Clock Drift and Y2K38 Failures}

\author{
Saeid~Jamshidi,
Foutse~Khomh,
Carol~Fung,
Omar~Abdul~Wahab,
and Rolando~Herrero
\thanks{Corresponding author: S. Jamshidi 
(\texttt{saeid.jamshidi@polymtl.ca}).}%
\thanks{S. Jamshidi and F. Khomh are with the SWAT Laboratory, 
Polytechnique Montréal, Québec, Canada 
(emails: \texttt{saeid.jamshidi@polymtl.ca}; 
\texttt{foutse.khomh@polymtl.ca}).}%
\thanks{O. Abdul~Wahab is with the Department of Computer Engineering 
and Software Engineering, Polytechnique Montréal, Québec, Canada 
(email: \texttt{omar.abdul-wahab@polymtl.ca}).}%
\thanks{C. Fung is with CIISE, Concordia University, Montréal, QC, Canada 
(email: \texttt{carol.fung@concordia.ca}).}%
\thanks{R. Herrero is with Northeastern University, Boston, MA, USA 
(email: \texttt{r.herrero@northeastern.edu}).}%
}

\maketitle
\begin{abstract}
Time integrity across distributed Internet of Things (IoT) devices is fundamental to reliable sensing, control, and security in \emph{energy cyber-physical systems}, including smart grids, microgrids, and distributed energy management platforms. However, operational energy IoT systems remain vulnerable to clock-drift escalation, time-synchronization manipulation, and catastrophic timestamp discontinuities, e.g., the Year~2038 (Y2K38) Unix epoch overflow. These failures violate timestamp monotonicity, distort temporal ordering, and introduce structured inconsistencies in system observations. Conventional anomaly detection models, which typically assume reliable and uniformly ordered timestamps, are therefore ill-equipped to capture timing-layer failures. This paper introduces \emph{STGAT} (\textbf{Spatio-Temporal Graph Attention Network}), a clock-dynamics-aware anomaly detection solution that jointly models temporal distortion and inter-device consistency in energy IoT systems. STGAT integrates drift-aware temporal embeddings and temporal self-attention to capture non-uniform and corrupted time evolution within individual device streams, while graph attention models the spatial propagation of timing inconsistencies across interconnected nodes. A Jacobian-regularized latent representation further promotes geometric separation between nominal clock evolution and anomalous temporal deformation caused by drift escalation, synchronization offsets, jitter accumulation, and epoch-overflow events. Experimental evaluation on energy IoT telemetry augmented with controlled timing-layer perturbations shows that the final STGAT model achieves 95.7\% accuracy, 94.0\% precision, 92.0\% recall, 93.0\% F1-score, and 0.97 AUC under the primary test setting. STGAT also reduces detection delay to 2.3 time steps, corresponding to a 26\% improvement over the closest baseline, while maintaining stable performance under overflow-induced discontinuities, stealthy drift escalation, and temporally induced physical inconsistencies.
\end{abstract}

\begin{IEEEkeywords}
Internet of Things (IoT), Time Synchronization, Y2K38, Timestamp Drift, NTP Spoofing, GPS Spoofing, Temporal Anomaly Detection, Spatio-Temporal Graph Attention Network (STGAT)
\end{IEEEkeywords}

\section{Introduction}
\label{Introduction}
The integration of Internet of Things (IoT) devices into modern energy cyber-physical systems, including smart grids, microgrids, distributed energy resources, and electric vehicle charging infrastructure, has transformed monitoring, protection, and control~\cite{smartgrid_iot2021, iot_energy2021, ev_charging2025}. These systems operate as tightly coupled spatio-temporal processes in which sensing, actuation, state estimation, protection, and event correlation depend on the correct ordering and alignment of distributed timestamps. Consequently, \emph{temporal integrity}, defined as the correctness, monotonicity, and cross-device consistency of time, is a core system invariant rather than a low-level implementation detail.
Violations of temporal integrity can disrupt multiple system layers. At the sensing layer, corrupted timestamps distort data alignment and degrade fusion~\cite{finkenzeller2025sensor}. At the control layer, timing errors can impact feedback loops, coordination mechanisms, and time-dependent decisions~\cite{liu2025smartgrid_sync,hassan2026sampled}. At the security layer, they can undermine anomaly detection, event correlation, and forensic analysis, all of which rely on temporally ordered observations~\cite{krishnamurthy2026tracking,breitinger2025sok}. These effects are particularly critical in applications such as phasor measurement units, distributed state estimation, coordinated protection, and latency-sensitive IoT media and audio systems, where accurate timing supports synchronization, buffering, event alignment, and real-time inference~\cite{balakrishnan2023clock,li2025high,jellum2022syncline,chen2025edge}.
Maintaining temporal integrity in distributed energy IoT systems is challenging because embedded devices rely on low-cost oscillators, synchronization protocols, and networked timing sources that are affected by environmental variability, aging, manufacturing tolerances, latency, jitter, and correction noise~\cite{drift_model2021,wang2025clock}. Common timing failures include gradual clock drift, abrupt synchronization offsets from NTP, PTP, and GNSS corrections, stochastic jitter from communication variability, adversarial synchronization manipulation, and catastrophic timestamp discontinuities such as epoch-overflow and Y2K38 events~\cite{yiugitler2020overview}. Such failures can desynchronize sensors, violate event ordering, delay and prematurely trigger control actions, and propagate across interconnected devices through shared synchronization and coordination processes~\cite{abdul2025deep,nasrullah2025seti,spanghero2024time,baccari2026cybersecurity}.
Despite these risks, existing anomaly detection approaches are poorly equipped to reason explicitly about timing-layer failures. Conventional time-series models often assume reliable timestamps and consistent temporal ordering~\cite{11494901}. Recurrent models rely on sequential consistency, transformer-based models depend on positional encoding, and graph-based models often assume synchronized inputs and stable temporal windows~\cite{lu2025survey,9338317,deng2021gdn,jin2024gnn4ts}. Even methods designed for irregular time series commonly treat timestamp irregularity as a sampling artifact handled through interpolation, resampling, and positional encoding, rather than as a temporal corruption process affecting distributed devices~\cite{chefrour2022evolution,wei2023coformer,liu2026irregular}. As a result, they remain limited when clock drift, synchronization faults, jitter accumulation, and timestamp discontinuities corrupt the temporal structure itself.
Spatio-temporal modeling offers a promising direction because it can jointly analyze temporal distortions and cross-device dependencies. To address this gap, this paper introduces \emph{STGAT} (Spatio-Temporal Graph Attention Network), a clock-dynamics-aware learning solution for detecting \emph{timing-layer failures} in energy IoT systems. STGAT models device-reported time as a deformable temporal signal governed by stochastic clock drift, synchronization offsets, jitter accumulation, and epoch-overflow events, and embeds these timing dynamics directly into the learning process through drift-aware temporal embeddings. It combines \emph{temporal self-attention}, which captures non-uniform and corrupted time evolution within individual device streams, with \emph{graph attention}, which models the spatial propagation and cross-device consistency of temporal distortions across interconnected nodes. To distinguish benign clock evolution from anomalous temporal behavior, a Jacobian-based geometric regularization term~\cite{ayaz2023improving} promotes the separation of nominal and corrupted temporal trajectories in the latent representation space. An online sequential detection mechanism further enables early identification of timing anomalies before temporal faults propagate into observable physical and operational inconsistencies.

The main contributions of this paper are summarized as follows:
\begin{enumerate}
  \item \textbf{Temporal integrity modeling for energy IoT:}
  We formalize timing-layer failure modes, including clock-drift escalation, synchronization-offset shocks, jitter accumulation, and epoch-overflow events, as deformations on temporal manifolds.

  \item \textbf{Jacobian-regularized spatio-temporal graph learning:}
  We propose STGAT, a clock-dynamics-aware spatio-temporal graph attention architecture that jointly models temporal deformation and inter-device dependencies through drift-aware embeddings, temporal self-attention, graph attention, and Jacobian-regularized latent geometry.

  \item \textbf{Empirical evaluation under timing failures:}
  We evaluate STGAT under diverse timing-layer failure scenarios and demonstrate improved detection accuracy, reduced detection latency, and statistically significant robustness compared to representative temporal, graph-based, and hybrid baselines.
\end{enumerate}

The paper is organized as follows. Section~\ref{sec:related} reviews related work. Section~\ref{sec:proposed_model} presents the proposed STGAT architecture, its timing-distortion operators, and its main learning components. Section~\ref{sec:dataset_preprocessing} describes the dataset, timing-perturbation generation, preprocessing steps, experimental settings, and runtime measurement protocols. Section~\ref{sec:experimental_findings} reports the experimental results and associated analyses. Section~\ref{sec:discussion} discusses the implications of the findings for temporal integrity and resilient energy IoT operation. Section~\ref{sec:limitations_future_work} outlines the study limitations and directions for future work. Section~\ref{sec:conclusion} concludes the paper.

\section{Related Work}
\label{sec:related}
This section reviews prior studies on time synchronization in smart grids and industrial IoT, adversarial time-synchronization attacks and spoofing-resilient defenses, and deep spatio-temporal anomaly detection models for multivariate time series.

\subsection{Time Synchronization in Smart Grids and Industrial IoT}
Accurate and stable time synchronization is foundational for large-scale IoT and energy cyber-physical systems. Surveys by Y{\i}gitler et al.~\cite{yigitler2020overview} and Liu et al.~\cite{liu2023clocksync} highlight trade-offs in accuracy, energy, and hardware complexity and review oscillator behavior, protocol-level mechanisms, and delay-sensitive cross-layer design. Liu et al.~\cite{liu2025smartgrid_sync} further emphasize strict timing requirements in PMUs, protection relays, and AMI, noting reliance on GNSS- and PTP-based synchronization. These studies underscore the criticality of timing but largely assume nominal operating conditions.

\subsection{Time-Synchronization Attacks and Spoofing Countermeasures}
Adversarial manipulation of time references, particularly via GNSS/GPS spoofing, has been widely studied. Zhang et al.~\cite{zhang2020gps_tsa} analyze GPS spoofing in power systems and review hardware and signal-level defenses. Xue et al.~\cite{xue2021gps_pmu} and Khan et al.~\cite{khan2022incremental_gps} propose data-driven detection and incremental spoofing strategies. Deep learning approaches include LSTM-based detection~\cite{liu2024tsa_lstm}, recurrent models for measurement correction~\cite{sabouri2023dl_gps}, and multi-source signal validation~\cite{ou2025tsa_protection}. These approaches are effective in specific scenarios but generally do not model system-wide temporal deformation.

\subsection{Deep and Spatio-Temporal Graph Models for Time-Series Anomaly Detection}
Deep learning dominates anomaly detection in multivariate time series for industrial and IoT systems~\cite{jia2025deep_ts_survey}. Hybrid signal-processing and DL pipelines improve robustness~\cite{backhus2025ts_sp_dl}, while transformer-based models such as RTdetector leverage global attention to model complex temporal anomalies~\cite{liu2025rtdetector}. Spatio-temporal graph neural networks (STGNNs) capture dependencies among interconnected devices. GST-Pro~\cite{zheng2024gstpro} uses graph-based controlled differential equations; Di et al.~\cite{di2024gnn_satellite} propose interpretable GNNs for satellite power systems; and Lu et al.~\cite{lu2024stgnn_iot} demonstrate the benefits of STGNNs in industrial IoT. AlZahrani et al.~\cite{alzahrani2025iot_stream} discuss real-time streaming challenges, concept drift, and resource efficiency.

Existing spatio-temporal graph neural networks and transformer-based anomaly detection methods can capture temporal and spatial dependencies when timestamps are reliable, but they do not explicitly model timing-layer corruption caused by clock drift, synchronization offsets, jitter accumulation, and epoch-overflow events. Even models designed for irregular sampling and temporal warping typically treat timestamp irregularity as a representation and sampling issue while still assuming monotonically ordered time. In contrast, STGAT treats time corruption as a first-class phenomenon by modeling device-reported timestamps as deformable signals governed by clock dynamics and overflow behavior. By combining clock-aware temporal features with graph attention, STGAT captures both local timestamp deformation and the spatial propagation of timing inconsistencies across interconnected energy IoT devices. This enables principled detection of temporal-integrity failures, including Y2K38-induced disruptions, that methods that assume timestamp reliability may miss.

\section{Proposed Model}
\label{sec:proposed_model}
This section presents STGAT, a Jacobian-regularized spatio-temporal graph attention network for detecting \emph{timing-layer failures} in energy IoT systems. STGAT treats device-reported timestamps as deformable temporal signals subject to clock drift, synchronization offsets, jitter accumulation, and epoch-overflow events, rather than assuming reliable, uniformly ordered time. This allows timing distortions to be modeled directly at the timestamp level rather than treated as noise and sampling artifacts.
At its core, STGAT integrates three components: drift-aware temporal embeddings, transformer-based temporal self-attention, and multi-head graph attention. The drift-aware embedding layer encodes timing variables associated with inter-sample spacing, cumulative drift, synchronization offsets, jitter, and overflow indicators. Temporal self-attention then captures non-uniform and corrupted time evolution within each device stream, enabling detection of weak but persistent drift patterns and abrupt offset-induced discontinuities from temporal context. In parallel, graph attention models communication and functional dependencies among IoT devices, allowing timing inconsistencies observed at one node to inform anomaly inference at related nodes. STGAT's design is guided by three principles. First, time is represented as a deformable signal rather than a fixed index, enabling explicit modeling of timestamp-monotonicity violations, reversed ordering, discontinuities, and non-causal timestamp sequences. Second, we jointly model temporal and spatial dependencies, enabling local timing anomalies to be interpreted in relation to cross-device propagation patterns. Third, physics-informed temporal operators and Jacobian-regularized latent representations distinguish smooth nominal clock evolution from anomalous temporal deformation. Jacobian-regularized latent modeling further improves the separation between benign and corrupted timing behavior. Under nominal conditions, clock evolution is expected to follow smooth and locally consistent latent trajectories. In contrast, drift escalation, synchronization-offset shocks, jitter bursts, and Y2K38-induced timestamp discontinuities introduce irregular geometric deviations. By regularizing local latent deformation, STGAT improves the discriminability of timing-layer anomalies in complex temporal graph settings.
The novelty of STGAT lies in the domain-specific integration of time-aware embeddings, physics-informed drift and overflow operators, graph-based propagation modeling, and Jacobian-regularized latent geometry for timing-integrity monitoring in energy IoT systems. Unlike standard recurrent, transformer-based, and graph-based anomaly detectors, STGAT explicitly targets temporal integrity violations caused by clock deformation, synchronization faults, and epoch-overflow behavior.
\begin{figure*}[!t]
    \centering
    \includegraphics[width=0.70\textwidth]{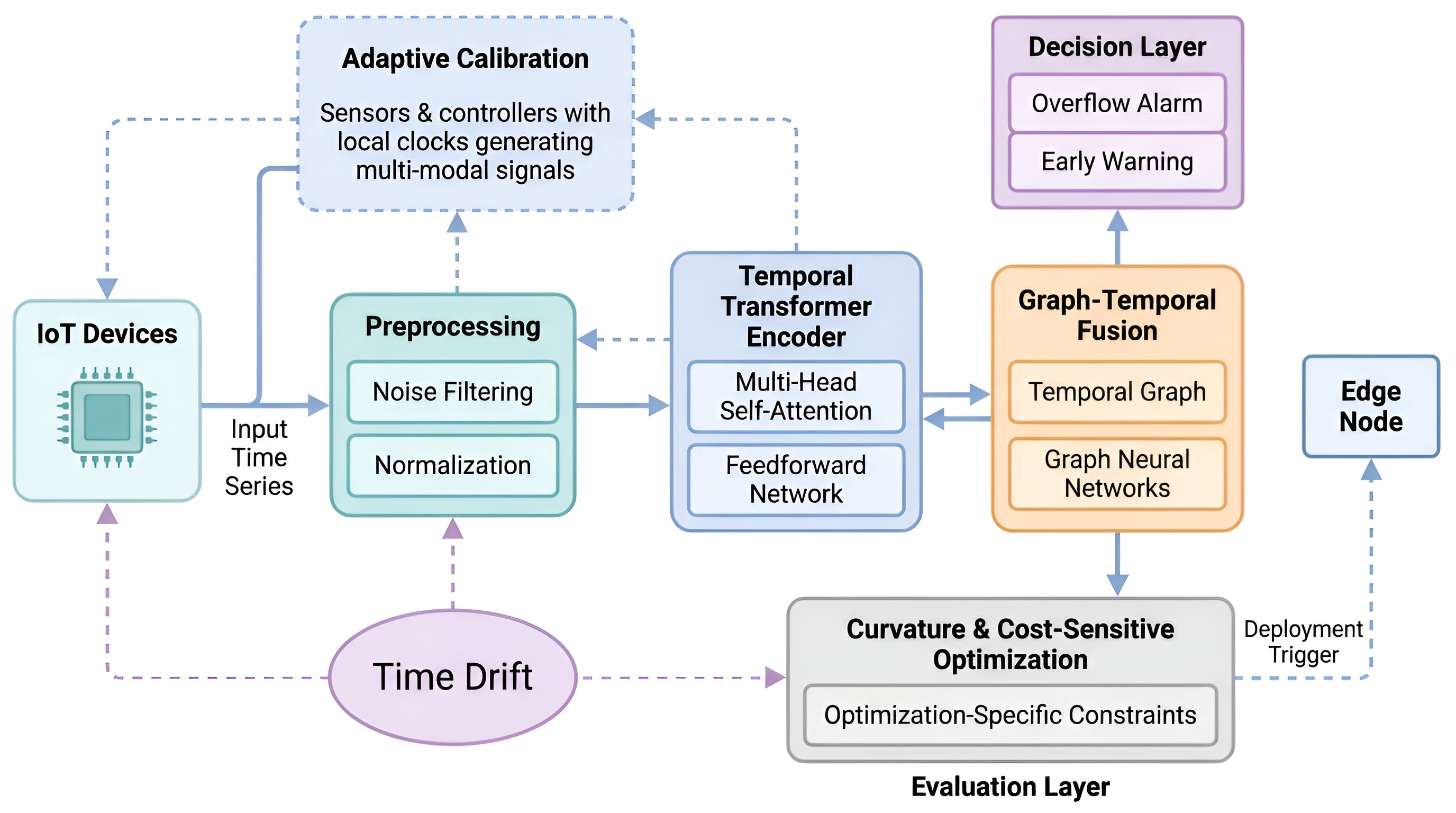} 
    \caption{Overview of the proposed STGAT architecture.}
    \label{fig:architecture_overview}
\end{figure*}
Figure~\ref{fig:architecture_overview} depicts the STGAT pipeline. Multivariate time series are augmented with drift-aware temporal features, encoded through temporal self-attention, and fused through graph attention to capture inter-device dependencies. A Jacobian-regularized decision layer then supports anomaly detection by separating nominal clock behavior from drift-, offset-, jitter-, and overflow-induced temporal deformation. This design enables robust timing-layer anomaly detection without requiring downstream physical anomalies to become visible before detection.

\subsection{System Operators and Distortion Modeling}
Each device $v_i$ produces a multivariate measurement stream:
\[
\mathbf{x}_i(t) \in \mathbb{R}^{F_s},\qquad 
t \in \{1,\dots,T\},
\]
where $F_s$ denotes the number of sensing modalities (e.g., voltage, current, frequency, and thermal measurements). We assume that all devices observe the underlying physical process synchronously in real time, but \emph{report} their measurements asynchronously due to clock drift, synchronization offsets, and potential epoch-overflow events. Importantly, temporal distortion affects the \emph{timestamp layer}, not the physical sensing process itself.
\paragraph*{Timestamp Operator}
Consider an energy IoT system composed of $N$ devices, indexed by $i \in \{1, \dots, N\}$. 
Each device $v_i$ maintains a local notion of time, represented as:
\[
\tau_i(t) = t + \delta_i(t) + \eta_i(t) + o_i(t)\,T_0 .
\]
Here, the device-reported timestamp $\tau_i(t)$ is decomposed into four components:
\begin{itemize}
    \item $t$: the true global physical time shared across all devices.
    \item $\delta_i(t)$: a continuous drift function (phase deviation), typically modeled as a stochastic process. It captures the cumulative frequency mismatch between the local oscillator of device $v_i$ and the global clock.
    \item $\eta_i(t)$: a discrete offset term representing abrupt time adjustments. These arise from GPS/NTP synchronization events and manual corrections applied at device $v_i$.
    \item $o_i(t)\,T_0$: an overflow term activated when the device's internal 32-bit signed Unix timestamp counter exceeds $T_0 = 2^{31}$. This explicitly models the Year~2038 (Y2K38) rollover.
\end{itemize}
The overflow term is critical: once a rollover occurs, timestamps wrap into the negative interval $[-2^{31}, -1]$, violating temporal monotonicity. 
Our formulation is firmware- and hardware-agnostic, modeling overflow as a fixed, device-local temporal discontinuity of magnitude $T_0$. We isolate temporal distortion as:
\[
\Psi_i(t) = \tau_i(t) - t,
\qquad 
\Delta\Psi_i(t) = \Psi_i(t+\Delta t) - \Psi_i(t).
\]
Here,
\begin{itemize}
    \item $\Psi_i(t)$ captures the absolute deviation between local device time and true physical time.
    \item $\Delta\Psi_i(t)$ quantifies the \emph{incremental distortion}, which is central to anomaly detection because it encodes whether the device clock advances faster or slower than the physical process.
\end{itemize}
Under nominal conditions, $\Delta\Psi_i(t) \approx 0$. In contrast, drift escalation, spoofing-induced offset shocks, and Y2K38 rollover events produce pronounced deviations, with epoch overflow introducing a catastrophic discontinuity of magnitude $T_0$. Moreover, we define a drift-aware sampling operator:
\[
(\mathcal{S}_i \mathbf{x})(t) = \mathbf{x}\big(t + \Psi_i(t)\big),
\]
which reflects that devices sense the physical world at the correct physical time, whereas the \emph{timestamps associated with the recorded samples} are distorted. As a result, the dataset is indexed by the corrupted time $\tau$ rather than the true time $t$. From the perspective of the learning algorithm, the observed data lie on a warped temporal manifold:
\[
\mathcal{M}_i = \{ (t, \mathbf{x}_i(t)) \mapsto (\tau_i(t), \mathbf{x}_i(t)) \}.
\]
This temporal warping is fundamental to anomaly detection: misaligned timestamps introduce false causality, cause identical events across devices to appear at different temporal indices, and lead attention-based architectures to compute correlations over inconsistent time references. Y2K38 rollover injects an extreme discontinuity of magnitude $T_0$, which can severely destabilize temporal reasoning unless explicitly modeled. This operator-level formulation enables principled discrimination among benign clock drift, synchronization-induced offset jumps, overflow-driven discontinuities, and intentional violations of temporal monotonicity. The remainder of the proposed methodology builds directly upon these operators, constructing drift-aware temporal embeddings, spatio-temporal inference mechanisms, and Jacobian-regularized optimization strategies that maintain STGAT's robustness under severe timestamp distortions, including Y2K38 rollover events.

\subsection{Algorithm 1: Drift-Aware Dataset Construction}
Algorithm~\ref{algorithm_1} constructs a physically grounded dataset that exposes the model to a wide range of temporal failure modes in energy IoT systems. These include oscillator drift, synchronization offset jumps, desynchronization shocks, adversarial timestamp perturbations, and the Y2K38 epoch overflow. Rather than serving as a simple data-augmentation routine, this algorithm explicitly simulates device-level time deformation and constructs warped temporal manifolds. We then use these manifolds to train the STGAT architecture.

\paragraph*{Drift Evolution} 
The oscillator-induced timestamp divergence is modeled via an Ornstein-Uhlenbeck \cite{higashida2026analysis} process discretized with Euler-Maruyama:
\[
\delta_{i,t} = \alpha_i \delta_{i,t-1} + \sigma_i \sqrt{\Delta t}\,\epsilon_t.
\]
This captures gradual deviations of device clocks from the global reference.

\paragraph*{Synchronization Offsets} 
Abrupt adjustments due to NTP corrections and PPS glitches are encoded as:
\[
\eta_{i,t} = \eta_{i,t-1} + \xi_t.
\]
Timing jitter is represented separately by $\jmath_{i,t}$ as a short-term variation in timestamp progression and is incorporated explicitly into the timing-distortion vector.

\paragraph*{Overflow Handling} 
The catastrophic Y2K38 rollover is modeled with:
\[
o_{i,t} = \mathbb{I}(\tau_{i,t-1} \ge T_0), \qquad T_0 = 2^{31}.
\]

\paragraph*{Timestamp Composition} 
The final distorted timestamp for device $v_i$ is:
\[
\tau_{i,t} = t + \delta_{i,t} + \eta_{i,t} + o_{i,t} T_0, 
\qquad 
\Psi_i(t) = \tau_{i,t} - t.
\]
Here, $\Psi_i(t)$ denotes the absolute deviation between the reported and true times.

\paragraph*{Drift-Aware Embedding}
Temporal irregularities are encoded in the timing-distortion vector:
\[
\mathbf{d}_t =
[\Delta t_t,\delta_t,\eta_t,\jmath_t,o_t]^\top
\in \mathbb{R}^{d_d},
\qquad d_d=5,
\]
where $\Delta t_t$ denotes the inter-sample interval, $\delta_t$ represents cumulative clock drift, $\eta_t$ denotes the synchronization offset, $\jmath_t$ represents timing jitter, and $o_t$ denotes the epoch-overflow indicator.
The timing-distortion vector is incorporated into the timing-aware feature representation:
\[
\mathbf{z}_t = \mathbf{x}_t + W_t\mathbf{d}_t .
\]
Here, $W_t$ projects the five-dimensional timing-distortion vector into the measurement-feature space, enabling the model to jointly represent observed telemetry and timing-layer behavior.

\paragraph*{Jacobian-Based Geometric Distortion}
The local geometric distortion of the timing-aware representation is approximated as:
\[
K(t)=\left\|J_t^\top J_t-I_{d_d}\right\|_F,
\]
where
\[
J_t=
\frac{\partial \mathbf{z}_t}
{\partial \mathbf{d}_t}
\in
\mathbb{R}^{d_z\times d_d},
\qquad
{d_d=5}.
\]
Here, $J_t$ denotes the timing Jacobian and is distinct from the jitter variable $\jmath_t$. The quantity $K(t)$ is a Jacobian-based proxy for local geometric distortion and should not be interpreted as the full Riemannian curvature \cite{lee2006riemannian} of the latent space. Rather, it measures how much the local timing-conditioned transformation departs from an approximately geometry-preserving mapping with respect to the five timing-distortion variables.
Under nominal timing conditions, $K(t)$ is expected to remain relatively small, whereas drift escalation, abrupt synchronization offsets, jitter bursts, and Y2K38-induced timestamp discontinuities may produce larger values.

\paragraph*{Dataset Tuple} 
Each sample is represented as $(\mathbf{z}_t, \Psi_i(t), K(t))$, encoding:
\begin{itemize}
    \item The warped signal representation,
    \item The underlying timestamp distortion, and
    \item The local latent manifold geometry.
\end{itemize}
This compact characterization allows the model to learn Y2K38-aware detection patterns.
\begin{algorithm}[H]
\footnotesize
\caption{Drift-Aware Data Construction}
\label{algorithm_1}
\begin{algorithmic}[1]

\State $\mathcal{D} \gets \emptyset$

\For{$i=1$ to $N$}
  \State Initialize $\delta_{i,0}, \eta_{i,0}, o_{i,0} \gets 0$

  \For{$t=1$ to $T$}

    \State $\delta_{i,t} \gets \alpha_i \delta_{i,t-1}
    + \sigma_i \sqrt{\Delta t}\,\epsilon_t$

    \State $\eta_{i,t} \gets \eta_{i,t-1} + \xi_t$

    \State $o_{i,t} \gets
    \mathbb{I}(\tau_{i,t-1} \ge T_0)$

    \State $\tau_{i,t} \gets
    t + \delta_{i,t} + \eta_{i,t} + o_{i,t}T_0$

    \State Obtain $\jmath_t$ from the corresponding timing-jitter feature

    \State $\mathbf{d}_t \gets
    [\Delta t_t,\delta_t,\eta_t,\jmath_t,o_t]^\top$

    \State $\mathbf{z}_t \gets
    \mathbf{x}_t + W_t\mathbf{d}_t$

    \State $
    K(t) \gets
    \left\|J_t^\top J_t-I_{d_d}\right\|_F
    $

    \State Insert $(\mathbf{z}_t,\Psi_i(t),K(t))$
    into $\mathcal{D}$

  \EndFor
\EndFor

\State \Return $\mathcal{D}$

\end{algorithmic}
\end{algorithm}

\subsection{Transformer, Graph Fusion, and Optimization Operators}
STGAT integrates drift-aware temporal embeddings, transformer-based self-attention, graph-based spatial fusion, and clock-dynamics-informed regularization into a unified learning framework. These components jointly capture non-uniform temporal sampling, timestamp distortion, inter-device dependencies, and epoch-overflow dynamics in energy IoT systems. The physics-informed aspect of STGAT is restricted to device-clock behavior and timestamp-generation mechanisms; the model does not impose an explicit physical model of the underlying energy system. Let the timing-distortion vector be defined as
\[
\mathbf{d}_t =
[\Delta t_t,\delta_t,\eta_t,\jmath_t,o_t]^\top
\in \mathbb{R}^{d_d},
\qquad d_d=5,
\]
where $\Delta t_t$ denotes the observed inter-sample interval, $\delta_t$ represents cumulative clock drift, $\eta_t$ denotes the synchronization offset, $\jmath_t$ captures short-term timing jitter, and $o_t$ indicates an epoch-overflow event. The corresponding timing-aware embedding is
\[
\mathbf{z}_t
=
\mathbf{x}_t + W_t\mathbf{d}_t,
\]
where $W_t$ projects the five-dimensional timing-distortion vector into the measurement-feature space, enabling its direct fusion with the observed feature vector $\mathbf{x}_t$. This representation explicitly incorporates inter-sample spacing, cumulative clock drift, synchronization offsets, timing jitter, and overflow indicators into the learned feature space. Under nominal operation, these timing variables evolve smoothly and produce relatively small changes in the timing-conditioned representation. In contrast, drift escalation, synchronization-offset shocks, jitter bursts, and Y2K38-induced timestamp rollover produce stronger and potentially discontinuous changes in the learned temporal representation.

\paragraph*{Temporal Self-Attention}
Temporal dependencies are modeled using the self-attention operator
\[
\mathcal{A}(Z)
=
\mathrm{softmax}\!\left(
\frac{ZW^Q(ZW^K)^\top}{\sqrt{d}}
\right)ZW^V,
\]
where $W^Q$, $W^K$, and $W^V$ denote the query, key, and value projection matrices, respectively.
Because the input representations contain timestamp-derived information, the resulting attention weights are conditioned on both the observed measurements and their temporal integrity. Under nominal operation, temporally adjacent and physically related observations are expected to exhibit relatively coherent attention patterns. In contrast, drift escalation, offset shocks, jitter accumulation, and epoch-overflow discontinuities disrupt this temporal consistency and alter the learned attention structure.

\paragraph*{Graph-Based Spatial Fusion}
Spatial dependencies among devices are incorporated through graph attention:
\[
h_i^{(\ell+1)}
=
\sigma\!\left(
\sum_{j\in\mathcal{N}(i)}
\alpha_{ij}Wh_j^{(\ell)}
\right),
\]
with
\[
\alpha_{ij}
=
\frac{
\exp\!\left(
a^\top
[
Wh_i^{(\ell)}
\Vert
Wh_j^{(\ell)}
]
\right)
}{
\sum_{k\in\mathcal{N}(i)}
\exp\!\left(
a^\top
[
Wh_i^{(\ell)}
\Vert
Wh_k^{(\ell)}
]
\right)
}.
\]
This operator lets each device's anomaly representation incorporate timing evidence from topologically or functionally related devices. Consequently, STGAT can identify correlated temporal-integrity violations, including synchronization failures and overflow-related inconsistencies affecting multiple devices that share communication paths, timing sources, or timestamp logic.

\paragraph*{Optimization Objective}
Model training is guided by the composite objective
\begin{align}
\mathcal{L}
=
&\;
\lambda_{\mathrm{rec}}
\|X-\hat{X}\|_2^2
+
\lambda_{\mathrm{cls}}\,
\mathrm{CE}(y,\hat{p})
\notag\\
&
+
\lambda_{\delta}
\sum_t
\left\|
\nabla\hat{\delta}_t
-
\nabla\delta_t
\right\|_1
\notag\\
&
+
\lambda_K
\sum_t
\left(
K(t)-\mu_K
\right)_+^2 ,
\end{align}
where $(u)_+=\max(u,0)$. The objective jointly enforces reconstruction fidelity, anomaly-classification accuracy, consistency with available clock-drift supervision, and Jacobian-based control of local geometric deformation.
The drift-consistency term is used only when an appropriate timing-supervision signal is available during training or calibration. It is not required for operational inference. In controlled supervised training, injected timing perturbations provide known drift-related quantities, and $\lambda_\delta>0$ regularizes the predicted drift increment toward the injected drift dynamics:
\[
\mathcal{L}_{\delta}^{\mathrm{sup}}
=
\sum_t
\left\|
\nabla\hat{\delta}_t
-
\nabla\delta_t
\right\|_1 .
\]
In semi-supervised calibration, exact drift trajectories can be replaced by observable timing proxies:
\[
\mathcal{L}_{\delta}^{\mathrm{proxy}}
=
\sum_t
\left\|
\nabla\hat{\delta}_t
-
\widetilde{\nabla\delta}_t
\right\|_1 ,
\]
where
\[
\widetilde{\nabla\delta}_{i,t}
=
\left[
\tau_i(t)-\tau_i(t-1)
\right]
-
\Delta t.
\]
The quantity $\widetilde{\nabla\delta}_{i,t}$ is a timestamp-progression proxy rather than a ground-truth drift label. It measures deviation from the expected local timestamp increment and can be refined with synchronization logs, heartbeat records, or external reference-clock measurements when available. During operational inference, no drift-consistency loss is evaluated, and no ground-truth drift trajectory is assumed. Equivalently, $\lambda_\delta=0$ with respect to online scoring. Detection instead uses the fixed trained model together with observed telemetry, reported timestamps, timestamp-derived features, temporal attention, graph fusion, and the learned anomaly decision outputs.
\begin{table}[!t]
\centering
\caption{Training and Deployment Settings for Drift-Consistency Learning}
\label{tab:deployment_settings}
\adjustbox{width=\columnwidth}{
\begin{tabular}{|l|c|c|p{4.0cm}|}
\hline
\textbf{Setting} &
\textbf{Timing Supervision} &
\textbf{$\lambda_\delta$} &
\textbf{Role of Drift-Consistency Term} \\
\hline

Controlled supervised training &
Injected drift quantities &
$>0$ &
Supervises predicted drift dynamics using known injected perturbations. \\
\hline

Semi-supervised calibration &
Timestamp-derived or auxiliary proxies &
$>0$ &
Uses partial timing supervision when exact drift labels are unavailable. \\
\hline

Operational inference &
Not required &
$0$ &
No drift-consistency loss is evaluated; anomaly decisions are generated by the fixed trained model from observed and timestamp-derived inputs. \\
\hline

\end{tabular}
}
\end{table}

\paragraph*{Jacobian-Based Geometric Regularization}
The quantity $K(t)$ is used as a Jacobian-based proxy for local geometric distortion and should not be interpreted as the Riemannian curvature of the latent manifold. Motivated by Jacobian regularization for controlling local model sensitivity and input--output deformation~\cite{wu2024adversarially}, let
\[
\mathbf{z}_t
=
\mathcal{E}_{\Theta}
(\mathbf{x}_t,\mathbf{d}_t),
\]
where
\[
\mathbf{d}_t
=
[\Delta t_t,\delta_t,\eta_t,\jmath_t,o_t]^\top
\in \mathbb{R}^{d_d},
\qquad d_d=5,
\]
with $\Delta t_t$ denoting the inter-sample interval, $\delta_t$ the cumulative clock drift, $\eta_t$ the synchronization offset, $\jmath_t$ the timing-jitter magnitude, and $o_t$ the epoch-overflow indicator.
The timing Jacobian is defined as
\[
J_t
=
\frac{\partial\mathbf{z}_t}
{\partial\mathbf{d}_t}
\in
\mathbb{R}^{d_z\times d_d},
\qquad d_d=5,
\]
and the corresponding geometric-distortion proxy is
\[
K(t)
=
\left\|
J_t^\top J_t-I_{d_d}
\right\|_F .
\]
This quantity measures the local sensitivity of the timing-conditioned representation to perturbations in the timing variables. Rather than estimating full manifold curvature, $K(t)$ measures the departure of the local timing-conditioned transformation from an approximately geometry-preserving mapping with respect to the five timing-distortion variables.
The Jacobian regularization term is defined as
\[
\mathcal{L}_{K}
=
\sum_t
\left(
K(t)-\mu_K
\right)_+^2,
\]
where $(u)_+=\max(u,0)$. This term penalizes only geometric deformation exceeding the tolerance $\mu_K$. Consequently, the regularizer limits excessive local sensitivity to timing perturbations while allowing the reconstruction and classification objectives to preserve discriminative information associated with genuine timing anomalies.
In implementation,
\[
D=
[\mathbf{d}_1,\ldots,\mathbf{d}_T]
\]
is marked as differentiable, and
\[
Z=\mathcal{E}_{\Theta}(X,D)
\]
is computed during the forward pass. We then obtain the timing Jacobian $J_t$ via automatic differentiation using vector-Jacobian or Jacobian-vector products. Because $d_d=5$, Jacobian computation is restricted to a low-dimensional timing input. The resulting additional training cost can be approximated as
\[
C_{\mathrm{J}}
\approx
\mathcal{O}
\left(
BTd_dC_{\mathrm{VJP}}
\right),
\qquad d_d=5,
\]
where $B$ denotes the mini-batch size, $T$ denotes the temporal-window length, and $C_{\mathrm{VJP}}$ denotes the cost of a vector-Jacobian product through the embedding module.
The Jacobian calculation and $\mathcal{L}_{K}$ are training-time operations and are not required for routine operational inference. Thus, the additional Jacobian-related overhead affects training but does not change the asymptotic inference cost of the temporal-attention and graph-attention modules. During deployment, the trained model parameters remain fixed, and online scoring uses the timing-aware embedding, temporal self-attention, graph fusion, and decision head. Per-window Jacobian computation is disabled during normal operational inference and is performed only when an explicit Jacobian-based diagnostic analysis is requested. Collectively, these operators provide a temporally grounded spatio-temporal learning framework for separating nominal timestamp evolution from drift-, synchronization-, jitter-, and overflow-induced timing failures in energy IoT networks. Algorithm~\ref{algorithm_2222} summarizes the training procedure and explicitly distinguishes the Jacobian-regularized optimization stage from the operational online-detection procedure described subsequently.
\begin{algorithm}[H]
\footnotesize
\caption{STGAT Training with Jacobian-Based Geometric Regularization}
\label{algorithm_2222}
\begin{algorithmic}[1]

\State Initialize model parameters $\Theta$

\For{$k=1$ to $N_{\mathrm{epoch}}$}
    \For{each mini-batch $\mathcal{B}$}

        \State Mark timing-distortion features $D$ as differentiable
        \State $Z \gets \mathcal{E}_{\Theta}(X,D)$

        \State $J_t
        \gets
        \partial\mathbf{z}_t/
        \partial\mathbf{d}_t$
        using automatic differentiation

        \State $K(t)
        \gets
        \left\|
        J_t^\top J_t-I_{d_d}
        \right\|_F$

        \State $H^{(0)} \gets Z$

        \For{$\ell=1$ to $L$}
            \State
            $H^{(\ell)}
            \gets
            \mathcal{A}(H^{(\ell-1)})
            +
            H^{(\ell-1)}$
        \EndFor

        \State
        $G
        \gets
        \mathcal{GAT}(H^{(L)})$

        \State
        $
        (\hat{X},\hat{p},\hat{\delta})
        \gets
        \mathcal{F}_{\Theta}(Z,G)
        $

        \State
        $\mathcal{L}_{rec}
        \gets
        \|X-\hat{X}\|_2^2$

        \State
        $\mathcal{L}_{cls}
        \gets
        \mathrm{CE}(y,\hat{p})$

        \If{exact injected drift supervision is available}
            \State
            $\mathcal{L}_{\delta}
            \gets
            \sum_t
            \|\nabla\hat{\delta}_t-\nabla\delta_t\|_1$

        \ElsIf{timestamp-derived drift proxy is available}
            \State
            $\mathcal{L}_{\delta}
            \gets
            \sum_t
            \|\nabla\hat{\delta}_t-
            \widetilde{\nabla\delta}_t\|_1$

        \Else
            \State
            $\mathcal{L}_{\delta}\gets0$
        \EndIf

        \State
        $\mathcal{L}_{K}
        \gets
        \sum_t
        (K(t)-\mu_K)_+^2$

        \State
        $\mathcal{L}
        \gets
        \lambda_{rec}\mathcal{L}_{rec}
        +
        \lambda_{cls}\mathcal{L}_{cls}
        +
        \lambda_{\delta}\mathcal{L}_{\delta}
        +
        \lambda_K\mathcal{L}_{K}$

        \State Backpropagate all active training-loss terms jointly

        \State
        $\Theta
        \gets
        \Theta
        -
        \alpha_k
        \nabla_{\Theta}\mathcal{L}$

    \EndFor
\EndFor

\State $\Theta^\star \gets \Theta$
\State \Return $\Theta^\star$

\end{algorithmic}
\end{algorithm}

\subsection{Online Detection Theory}
The online detection module continuously monitors each device's temporal integrity in real time. At each time step $t$, the model produces an anomaly posterior:
\[
\hat{p}_t = f_\Theta(\mathbf{z}_{1:t}),
\]
representing the probability that a timing-layer anomaly has occurred given all observations up to time $t$. In operational inference, the input to $f_\Theta$ consists of observed telemetry, reported timestamps, and timestamp-derived variables, e.g., inter-sample interval, local timestamp progression, jitter estimate, synchronization offset estimate when available, and overflow-related clock indicators. The online detector does not assume access to ground-truth drift labels, synchronization logs, or external reference clocks. The drift-consistency loss used during supervised training and semi-supervised calibration is not required to compute $\hat{p}_t$ during deployment.
This posterior is converted into a log-likelihood ratio:
\[
\Lambda_t = \log \frac{\hat{p}_t}{1-\hat{p}_t},
\]
which quantifies the evidence supporting an anomaly versus normal behavior. To accumulate information over recent observations, the detector computes a sequential score over a sliding window of length $T$:
\[
S_t =
\sum_{i=\max(1,t-T+1)}^{t}\Lambda_i .
\]
This helps detect persistent but subtle temporal anomalies that may not be evident at a single time step. Because the variance of $S_t$ may change over time, especially near epoch rollovers, a dynamic threshold is used:
\[
\theta_t =
\theta_0 +
\gamma
\sqrt{\mathrm{Var}(S_{t-w:t})},
\]
where $\theta_0$ is a base threshold, $\gamma$ is a scaling factor, and $w$ is the window over which variance is estimated.
In addition to the sequential likelihood, the detector can use timestamp-derived consistency checks that do not require labeled drift trajectories. Specifically, the predicted drift increment is compared against a local drift proxy estimated from reported timestamp increments:
\[
\widehat{\nabla \delta}_t =
\hat{\delta}_t-\hat{\delta}_{t-1},
\qquad
\widetilde{\nabla \delta}_t =
\tau_t-\tau_{t-1}-\Delta t_t ,
\]
\[
C_\delta(t)=
\left\|
\widehat{\nabla \delta}_t
-
\widetilde{\nabla \delta}_t
\right\| .
\]
Here, $\widetilde{\nabla \delta}_t$ is not a ground-truth drift label; it is an observable proxy derived from local timestamp progression. A cumulative timestamp-derived drift estimate can be updated as
\[
\widetilde{\delta}_t
=
\widetilde{\delta}_{t-1}
+
\widetilde{\nabla \delta}_t,
\qquad
\widetilde{\delta}_0=0.
\]
If synchronization logs, heartbeat records, or reference-clock measurements are available, they may refine the corresponding timing estimates. If they are unavailable, the detector still performs online scoring using timestamp-derived features, learned embeddings, graph propagation, sequential evidence, and the anomaly posterior.
Accordingly, the operational timing vector is represented as
\[
\mathbf{d}_t^{\mathrm{obs}}
=
[\Delta t_t,
\widetilde{\delta}_t,
\widetilde{\eta}_t,
\jmath_t,
o_t]^\top
\in\mathbb{R}^{d_d},
\qquad d_d=5,
\]
where $\widetilde{\eta}_t$ denotes the available synchronization-offset estimate, $\jmath_t$ denotes the timing-jitter estimate, and $o_t$ denotes the overflow-related indicator. When an auxiliary synchronization-offset estimate is unavailable, we obtain the corresponding feature from the timestamp-derived preprocessing used for operational inference.
The online decision can therefore combine sequential evidence, timestamp-proxy consistency, and overflow risk:
\[
\mathbb{I}\!\left(S_t>\theta_t\right), \quad
\mathbb{I}\!\left(C_\delta(t)>\epsilon_\delta\right), \quad
\mathbb{I}\!\left(P_{\mathrm{over}}(t)>\epsilon_o\right).
\]
These indicators capture: 1) whether the accumulated log-likelihood exceeds the adaptive threshold, 2)whether the predicted drift increment deviates from the timestamp-derived local drift proxy, and 3) whether the risk of a Y2K38-related overflow-and-rollover discontinuity is high. By combining sequential evidence with timestamp-derived supporting checks, the online detector remains robust even when individual cues are noisy or partially unavailable.
\subsection{Online Y2K38 Overflow and Drift Detection}
Algorithm~\ref{algorithm_2} summarizes the proposed online detection mechanism. The algorithm operates in the operational inference setting: it requires no ground-truth drift labels, synchronization logs, or reference clocks for online scoring. Instead, it integrates sequential likelihood accumulation, timestamp-derived drift-proxy consistency, timing-jitter information, and predicted overflow risk. This enables early identification of Y2K38-related timing failures and stealthy clock-manipulation events while remaining applicable to deployments without full timing supervision.
\begin{algorithm}[H]
\footnotesize
\caption{Online Drift and Overflow Detection without Ground-Truth Drift Labels}
\label{algorithm_2}
\begin{algorithmic}[1]

\State Initialize $\hat{\delta}_0 \gets 0$
\State Initialize $\widetilde{\delta}_0 \gets 0$

\For{each time step $t$}

    \State $\widetilde{\nabla\delta}_t
    \gets
    \tau_t-\tau_{t-1}-\Delta t_t$

    \State $\widetilde{\delta}_t
    \gets
    \widetilde{\delta}_{t-1}
    +
    \widetilde{\nabla\delta}_t$

    \State Obtain timing-jitter estimate $\jmath_t$

    \State Obtain synchronization-offset estimate $\widetilde{\eta}_t$ when available

    \State $\mathbf{d}_t^{\mathrm{obs}}
    \gets
    [\Delta t_t,
    \widetilde{\delta}_t,
    \widetilde{\eta}_t,
    \jmath_t,
    o_t]^\top$

    \State $\mathbf{z}_t
    \gets
    \mathcal{E}
    (\mathbf{x}_t,\mathbf{d}_t^{\mathrm{obs}})$

    \State $(\hat{p}_t,\hat{\delta}_t)
    \gets
    f_\Theta(\mathbf{z}_{1:t})$

    \State $\Lambda_t
    \gets
    \log\!\left(
    \hat{p}_t/(1-\hat{p}_t)
    \right)$

    \State $S_t
    \gets
    \sum_{i=\max(1,t-T+1)}^{t}\Lambda_i$

    \State $\theta_t
    \gets
    \theta_0+
    \gamma
    \sqrt{\mathrm{Var}(S_{t-w:t})}$

    \State $\widehat{\nabla\delta}_t
    \gets
    \hat{\delta}_t-\hat{\delta}_{t-1}$

    \State $C_\delta(t)
    \gets
    \left\|
    \widehat{\nabla\delta}_t
    -
    \widetilde{\nabla\delta}_t
    \right\|$

    \State $P_{\mathrm{over}}(t)
    \gets
    \sigma\!\left(
    w_o^\top
    [\hat{\delta}_t,v_t,a_t,o_t]
    \right)$

    \State $\mathcal{D}(t)
    \gets
    \mathbb{I}\!\left(
    S_t>\theta_t
    \;\wedge\;
    \left(
    C_\delta(t)>\epsilon_\delta
    \;\vee\;
    P_{\mathrm{over}}(t)>\epsilon_o
    \right)
    \right)$

\EndFor

\end{algorithmic}
\end{algorithm}

\subsection{Threat Model and Assumptions}
\label{sec:threat_model}
We consider an adversary that targets the \emph{temporal integrity} of energy IoT systems by inducing deviations between the true physical time $t$ and the device-reported timestamp $\tau_i(t)$. Such deviations violate temporal correctness, monotonicity, and cross-device consistency, disrupting sensing, control, coordination, logging, and forensic analysis, yet remain difficult to detect with conventional monitoring.
The threat model focuses on timing-layer attacks rather than direct manipulation of physical measurements. The adversary may introduce gradual clock-drift escalation, abrupt synchronization-offset shocks, jitter bursts, and epoch-overflow discontinuities. These attacks distort timestamps and event ordering while preserving the physical plausibility of sensor payloads, making them challenging for models that assume reliable timestamps and uniform sampling. We assume that a non-trivial subset of devices remains uncompromised and can provide temporal-spatial reference points. Adversarial perturbations are also bounded by realistic constraints on oscillators, synchronization protocols, firmware, and network latency. These assumptions define the operational scope for evaluating drift escalation, synchronization offsets, and Y2K38-induced discontinuities.

\paragraph*{Adversarial Capabilities}
The adversary can manipulate device-level timestamps by spoofing and delaying NTP messages, spoofing GPS/GNSS time references, and exploiting firmware-level clock-management vulnerabilities. The actions are modeled as perturbations applied to the timestamp operator:
\begin{equation}
\tau_i(t) = t + \delta_i(t) + \eta_i(t) + o_i(t)\,T_0 ,
\end{equation}
where $\delta_i(t)$ denotes gradual clock drift, $\eta_i(t)$ represents abrupt synchronization offset jumps, and $o_i(t)\,T_0$ encodes epoch-overflow discontinuities associated with the Y2K38 problem. Under attack conditions, the adversary seeks to maximize the temporal distortion:
\begin{equation}
\Psi_i(t) = \tau_i(t) - t
\end{equation}
while minimizing detectability, typically by inducing low-rate and piecewise-smooth perturbations that resemble benign clock evolution.

\paragraph*{Attack Objectives and Constraints}
The adversary is assumed to be constrained by physical and operational limits. In particular, it cannot arbitrarily modify raw physical measurements, e.g.,  voltage, at scale without triggering independent physical-layer alarms. Moreover, instantaneous, unbounded timestamp manipulation is infeasible due to oscillator dynamics, synchronization-protocol behavior, firmware constraints, and network latency. Consequently, feasible attacks satisfy:
\begin{equation}
\left| \Delta \Psi_i(t) \right| \leq \epsilon_t,
\qquad
\left| \nabla \delta_i(t) \right| \leq \epsilon_d,
\end{equation}
for small but cumulative bounds $\epsilon_t$ and $\epsilon_d$, incentivizing stealthy drift escalation and delayed offset injection rather than abrupt, easily detectable jumps.

\paragraph*{System Assumptions}
We assume that all devices observe the underlying physical process synchronously in real time, but report measurements using locally distorted clocks. The device-interaction graph $\mathcal{G} = (\mathcal{V}, \mathcal{E})$ is partially known and reconstructed from physical connectivity, logical dependencies, and observed communication metadata. At any given time, a non-trivial subset of devices remains uncompromised, enabling the exploitation of spatial and temporal consistency across nodes, i.e.,
\begin{equation}
\exists\, \mathcal{V}_{\text{clean}} \subset \mathcal{V}
\quad \text{s.t.} \quad
\Psi_i(t) \approx 0, \;\; \forall i \in \mathcal{V}_{\text{clean}} .
\end{equation}
This assumption is critical for leveraging graph-based attention to detect correlated timing inconsistencies.

\section{Dataset and Preprocessing}
\label{sec:dataset_preprocessing}
We evaluate STGAT using the publicly available \textit{Edge-IIoTset} dataset~\cite{ferrag2022edge}, which contains heterogeneous multi-device and multimodal telemetry collected from an IoT-enabled cyber-physical testbed. Edge-IIoTset provides the underlying sensing and communication observations used in our experiments; however, it was not originally designed to represent energy-specific timing-integrity failures, e.g., clock-drift escalation, synchronization-offset manipulation, jitter accumulation, and Y2K38-induced timestamp rollover. Its heterogeneous device streams, communication metadata, and temporally correlated observations nevertheless provide a suitable basis for controlled evaluation of timing-layer anomaly detection. Besides, we retain only features relevant to sensing, timing, communication behavior, and cyber-physical consistency. These include device-level measurements such as voltage, current, temperature, humidity, and power consumption, together with packet- and flow-level attributes, e.g., inter-arrival time, flow duration, packet length, protocol identifiers, and device identifiers. Moreover, we exclude features unrelated to temporal-integrity analysis. Device identifiers and available communication metadata are additionally used to construct the device-interaction graph employed by the graph-attention component of STGAT.
Because Edge-IIoTset does not contain native labels for the timing-layer failures targeted in this study, controlled temporal perturbations are introduced at the \emph{reporting layer}, while the underlying sensor payloads are preserved. Specifically, the reported timestamps are modified to emulate gradual clock drift, synchronization-offset shocks, timing jitter, and epoch-overflow behavior. Consequently, the injected timing anomalies manifest as distorted inter-sample intervals, cross-device temporal misalignment, event-ordering inconsistencies, and abrupt timestamp discontinuities rather than direct modification of the physical measurement values. The timing-anomaly labels used for the primary STGAT evaluation are therefore determined by the controlled perturbation schedule rather than inferred from the original Edge-IIoTset attack labels. Perturbations are applied to a subset of devices while the remaining devices preserve nominal timestamp progression, thereby providing temporally consistent reference nodes for graph-based comparison. Five time-aware variables are introduced, as summarized in Table~\ref{tab:time_features}: cumulative timestamp drift, drift rate, timing jitter, synchronization offset, and a Y2K38 overflow indicator. These variables are generated according to the timestamp-distortion operators defined in Section~\ref{sec:proposed_model}. The primary experiments correspond to the controlled supervised-training configuration summarized in Table~\ref{tab:deployment_settings}. During controlled training and validation, the injected perturbation process provides drift-, offset-, jitter-, and overflow-related supervision to evaluate whether STGAT learns the intended timing-failure mechanisms. These injected quantities are training/evaluation annotations and should not be interpreted as information required during operational deployment. Operational inference is explicitly separated from this controlled training setting. At deployment time, the trained detector does not require ground-truth drift trajectories, injected perturbation labels, synchronization logs, or an external reference clock. Instead, inference uses the fixed trained model together with observed telemetry, reported timestamps, timestamp-derived features, temporal attention, graph-based spatial dependencies, and the resulting anomaly scores. When exact drift supervision is unavailable, the drift-consistency loss is not evaluated during online scoring. Preprocessing consists of missing-value imputation, feature normalization, and segmentation into overlapping 60-sample temporal windows. To reduce device-level information leakage between training and evaluation partitions, the data are partitioned by device using a 70\%/10\%/20\% training, validation, and testing split, respectively. Thus, samples belonging to the same device are not arbitrarily distributed across the three partitions. Fixed random seeds are used throughout the experimental pipeline, and hyperparameters are selected using grid-search optimization on the validation partition. Accordingly, the resulting benchmark is best characterized as a timing-augmented version of Edge-IIoTset rather than a fully synthetic dataset: the underlying IoT telemetry originates from the original testbed, whereas the clock-drift, synchronization-offset, jitter, and Y2K38 perturbations are introduced under controlled experimental conditions. To complement this offline timing-augmented evaluation, STGAT is also assessed on the physical edge-gateway prototype described in Section~\ref{Edge}, with the corresponding runtime results reported in Section~\ref{sec:experimental_findings}. The physical testbed evaluates prototype-level runtime behavior under controlled timing perturbations and does not replace the offline benchmark as the primary model-training dataset.
\begin{table}[!t]
\centering
\caption{Time-Aware Features Added to the Edge-IIoTset Dataset}
\label{tab:time_features}
\setlength{\tabcolsep}{3pt}
\footnotesize
\resizebox{\columnwidth}{!}{
\begin{tabular}{|l|l|p{4.1cm}|}
\hline
\textbf{Feature} & \textbf{Type} & \textbf{Description} \\
\hline

timestamp\_drift &
Continuous &
Cumulative deviation of the device-reported timestamp from the nominal reference timeline. \\
\hline

drift\_rate &
Continuous &
Rate of change of cumulative timestamp drift, representing gradual clock-dynamics deviation. \\
\hline

jitter\_ms &
Continuous &
Short-term variation in successive timestamp intervals associated with timing and synchronization instability. \\
\hline

ntp\_offset\_ms &
Continuous &
Injected or estimated synchronization offset between device-reported time and the nominal timing reference. \\
\hline

epoch\_overflow\_flag &
Binary &
Indicator identifying activation of the modeled 32-bit Unix timestamp rollover associated with Y2K38. \\
\hline

\end{tabular}
}
\end{table}

\subsection{Edge Deployment Testbed}
\label{Edge}
This subsection presents a physical edge-computing prototype for evaluating the runtime behaviour of STGAT under the specified hardware configuration. Training and offline benchmarking use the augmented Edge-IIoTset dataset, whereas the edge testbed measures inference latency, energy consumption, CPU usage, and runtime behaviour under controlled timing perturbations that offline datasets alone cannot fully capture. The testbed emulates a small-scale energy IoT microgrid in which sensing devices generate telemetry on voltage, current, temperature, and power and transmit these measurements to an inference gateway via TCP messages that carry 32-bit Unix timestamps. The reporting layer injects controlled timing perturbations, including gradual clock drift, abrupt synchronization offsets, timing jitter, and Y2K38 epoch-overflow events, without modifying the underlying sensor values. This design ensures that the evaluated anomalies represent temporal-integrity violations rather than direct manipulation of the physical measurements. The edge testbed represents operational inference under a prototype configuration: the trained STGAT model processes telemetry, reported timestamps, and timestamp-derived features without requiring ground-truth drift labels, synchronization logs, heartbeat records, and an external reference clock. Accordingly, the experiment assesses runtime feasibility under the evaluated small-scale configuration rather than establishing general real-time guarantees. The supervised drift-consistency loss is not applied during online inference. Instead, anomaly scores are computed using timing-aware representations, temporal self-attention, graph attention, timestamp-derived drift proxies, overflow-risk estimates, and sequential likelihood accumulation, including when auxiliary timing information is incomplete. This setup enables the evaluation of 1) end-to-end inference latency, 2) CPU usage, 3) incremental energy consumption, 4) runtime behaviour under clock-drift and synchronization disturbances, and 5) execution stability under timing jitter and sensor noise. These measurements characterize prototype-level runtime behaviour under the evaluated hardware and workload conditions; they do not establish deadline compliance, tail-latency guarantees, sustained-load scalability, and performance under large concurrent-device populations.
\begin{figure}[!t]
    \centering
    \includegraphics[width=0.50\textwidth]{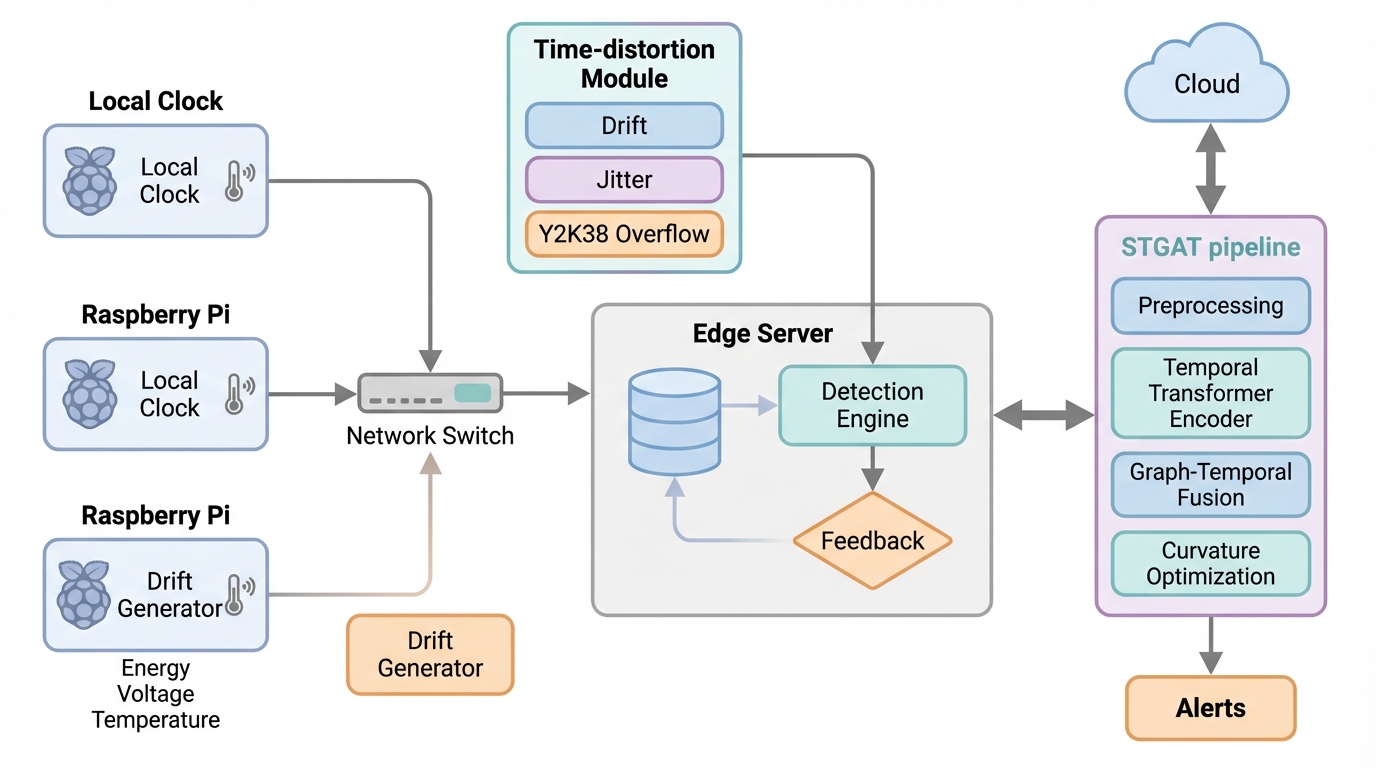}
    \caption{Schematic of the edge-gateway runtime evaluation setup.}
    \label{fig:testbed}
\end{figure}

\subsection{Edge Runtime Measurement Protocol}
\label{sec:edge_measurement_protocol}
STGAT was evaluated on the physical edge-gateway testbed shown in Figure~\ref{fig:testbed} using a fixed inference-only protocol. Model parameters remained unchanged throughout the evaluation, and we performed no online training, parameter adaptation, gradient computation, or Jacobian computation. Each inference processed a completed 60-sample overlapping window. Furthermore, we measured end-to-end inference latency from when the complete input window became available at the gateway until the corresponding anomaly decision was generated.
Besides, we collected measurements under five runtime conditions: nominal operation, drift escalation, synchronization offset, timing jitter, and Y2K38 overflow. Independent streaming trials used different perturbation schedules and message timings to capture variability across repeated executions. We discarded initial warm-up windows before calculating latency, CPU usage, power, and energy statistics. We measured CPU usage at the gateway-process level during active STGAT inference. We evaluated each runtime condition using 10 independent streaming trials. Each trial consisted of a 30-minute continuous inference stream and produced 300 completed inference windows after warm-up removal. Consequently, 3000 inference windows were evaluated per runtime condition, yielding 15\,000 windows across all five conditions. Also, we initially recorded runtime measurements at the individual-window level and then aggregated them within each independent trial. We treated the resulting 50 trial-level summaries, corresponding to 10 trials per runtime condition, as independent statistical observations. Reported descriptive statistics include the mean, standard deviation (SD), 95\% confidence interval (CI) of the mean, and minimum--maximum trial-level range. This trial-level aggregation avoids treating temporally correlated windows from the same continuous stream as independent experimental repetitions. Gateway power consumption was measured using a \textit{MakerHawk UM25C} USB power analyzer connected to the Raspberry Pi gateway and sampled at 1~Hz. Power measurements were collected under two configurations: 1) an idle streaming baseline, in which telemetry reception and window buffering remained active while STGAT inference was disabled, and 2) an active STGAT inference condition, in which the complete detection pipeline was executed. Moreover, we calculated relative power overhead and incremental energy consumption separately to distinguish power (watts) from energy (joules). The relative power overhead with respect to the idle streaming baseline was defined as
\[
\mathrm{Power\;Overhead}(\%) =
\frac{P_{\mathrm{STGAT}}-P_{\mathrm{baseline}}}
{P_{\mathrm{baseline}}}
\times 100,
\]
where $P_{\mathrm{STGAT}}$ and $P_{\mathrm{baseline}}$ denote the average gateway power consumption, measured in watts, during active STGAT inference and idle streaming, respectively.
Incremental energy consumption per inference was estimated as
\[
E_{\mathrm{inf}}
=
\left(
\overline{P}_{\mathrm{STGAT}}
-
\overline{P}_{\mathrm{baseline}}
\right)
\overline{T}_{\mathrm{inf}},
\]
where $\overline{P}_{\mathrm{STGAT}}$ and $\overline{P}_{\mathrm{baseline}}$ denote the trial-level average gateway power consumption during active STGAT inference and idle streaming, respectively, and $\overline{T}_{\mathrm{inf}}$ denotes the corresponding mean inference duration in seconds. Because the USB power analyzer samples at 1~Hz and cannot resolve individual inference pulses, $E_{\mathrm{inf}}$ represents an average incremental energy estimate per completed inference window. Since power is measured in watts and inference duration is measured in seconds, $E_{\mathrm{inf}}$ is expressed in joules. Thus, the relative power overhead characterizes the increase in average gateway power consumption during active inference, whereas incremental energy per inference characterizes the additional energy associated with processing one completed input window. The protocol provides repeated-trial estimates of average runtime behaviour and variability under the evaluated hardware, workload, and perturbation conditions. However, it does not define application-specific deadlines or evaluate latency percentiles, deterministic worst-case latency, maximum sustainable throughput, large concurrent-device populations, or multi-gateway scaling. Accordingly, the edge experiment is interpreted as a prototype-level assessment of runtime feasibility rather than evidence of general real-time deployability across heterogeneous operational energy IoT systems.
\begin{table*}[!t]
\centering
\footnotesize
\caption{Edge Runtime Measurement Configuration}
\label{tab:edge_runtime_config}

\begin{tabular}{|l|l|}
\hline
\textbf{Item} & \textbf{Configuration} \\
\hline
Inference device & Raspberry Pi 4 Model B edge gateway \\
\hline
Processor & Quad-core ARM Cortex-A72 64-bit CPU at 1.5 GHz \\
\hline
Memory & 4 GB LPDDR4 RAM \\
\hline
Operating system & Raspberry Pi OS 64-bit \\
\hline
Runtime environment & Python-based inference runtime with gradient computation disabled \\
\hline
Input-window size & 60 samples \\
\hline
Inference mode & Fixed trained model; no online training or parameter adaptation \\
\hline
Runtime Jacobian computation & Disabled during online inference \\
\hline
Runtime conditions & Nominal, drift escalation, synchronization offset, jitter, and Y2K38 overflow \\
\hline
Repeated trials & 10 independent streaming trials per runtime condition \\
\hline
Trial duration & 30 minutes per trial \\
\hline
Evaluated windows & 300 inference windows per trial after warm-up removal \\
\hline
Total windows & 3000 windows per condition; 15\,000 windows overall \\
\hline
Statistical unit & Trial-level summary ($n=50$ independent trials) \\
\hline
Warm-up policy & Initial warm-up windows discarded before measurement \\
\hline
Latency metric & End-to-end processing time per completed 60-sample window \\
\hline
CPU metric & Mean process-level CPU usage during active inference \\
\hline
Power measurement &
\textit{MakerHawk UM25C} USB power analyzer; 1-Hz sampling frequency \\
\hline
Power metric &
Relative increase in average gateway power consumption compared with the idle streaming baseline (\%) \\
\hline
Energy metric &
Incremental energy consumption per completed inference window (J) \\
\hline
Reported statistics & Mean, SD, 95\% CI of the mean, and trial-level range \\
\hline
Interpretation scope & Runtime feasibility under the evaluated prototype configuration \\
\hline
\end{tabular}

\end{table*}

\section{Experimental Findings}
\label{sec:experimental_findings}
This section presents the experimental evaluation of the proposed STGAT solution for detecting timing-layer anomalies and Y2K38-induced timestamp discontinuities in energy IoT systems.

\subsection{Performance Comparison}
\label{sec:performance_comparison}
We evaluated the proposed STGAT solution against generic baselines, including LSTM, Transformer, GAT, and IoT-TimeFormer, as well as representative spatio-temporal anomaly detection models such as MTAD-GAT, GDN, and Anomaly Transformer. Furthermore, we evaluated the models under timing-layer perturbations, including clock-drift escalation, synchronization-offset perturbations, jitter accumulation, and Y2K38-induced timestamp discontinuities. As shown in Table~\ref{tab:perf_baseline_comparison}, STGAT achieves the highest accuracy and AUC among the evaluated baselines. The broader comparison reported in Section~\ref{sec:previous_spatiotemporal_comparison} and Table~\ref{tab:prev_st_numerical_comparison} further shows that PHGAT achieves higher precision, recall, and F1-score. Therefore, we interpret the results metric-by-metric: STGAT's main advantage is high accuracy, strong AUC, and reduced detection delay under drift, offset, jitter, and Y2K38 perturbations, rather than uniform superiority across all metrics.
\begin{table}[!t]
\centering
\footnotesize
\setlength{\tabcolsep}{4pt}
\caption{Comparison of STGAT Performance Against Baseline and Spatio-Temporal Anomaly Detection Models}
\label{tab:perf_baseline_comparison}
\resizebox{\columnwidth}{!}{%
\begin{tabular}{|l|c|c|c|c|c|}
\hline
\textbf{Model} & \textbf{Accuracy (\%)} & \textbf{Precision (\%)} & \textbf{Recall (\%)} & \textbf{F1-score (\%)} & \textbf{AUC} \\
\hline
LSTM & 90.2 & 91.0 & 90.0 & 90.0 & 0.91 \\
Transformer & 92.1 & 92.0 & 90.0 & 91.0 & 0.92 \\
GAT & 91.0 & 90.0 & 90.0 & 90.0 & 0.91 \\
IoT-TimeFormer & 93.6 & 93.0 & 92.0 & 92.0 & 0.94 \\
MTAD-GAT & 93.0 & 91.8 & 91.0 & 91.8 & 0.94 \\
GDN & 92.7 & 91.5 & 91.0 & 91.5 & 0.94 \\
Anomaly Transformer & 93.2 & 92.1 & 91.5 & 92.1 & 0.94 \\
\textbf{STGAT (Proposed)} & \textbf{95.7} & \textbf{94.0} & \textbf{92.0} & \textbf{93.0} & \textbf{0.97} \\
\hline
\end{tabular}%
}
\end{table}
Unless stated otherwise, all full-model STGAT values refer to the same primary test setting: 95.7\% accuracy, 94.0\% precision, 92.0\% recall, 93.0\% F1-score, 0.97 AUC, and 2.3-step detection delay.
The results in Table~\ref{tab:perf_baseline_comparison} show that STGAT outperforms LSTM and Transformer, highlighting the limitations of purely temporal models when timestamps are unreliable. Its gain over GAT shows that spatial reasoning alone is insufficient when temporal ordering is corrupted. The improvement over IoT-TimeFormer further supports the benefit of combining drift-aware temporal embeddings with graph-based spatial fusion. Accordingly, STGAT is a timing-integrity-aware detector with advantages in high detection accuracy, high AUC, and early anomaly response to timing-layer perturbations, rather than a universally dominant model across all classification metrics.
Welch's $t$-tests~\cite{ahad2014sensitivity} show significant F1-score gains over the generic baselines ($p < 0.001$), with large Cohen's $d$ values: 1.90 for LSTM, 1.25 for Transformer, 1.40 for GAT, and 0.85 for IoT-TimeFormer.
\begin{table}[!t]
\centering
\caption{Pairwise Statistical Comparison Using Welch's $t$-Test on F1-score}
\label{tab:perf_welch_f1}
\setlength{\tabcolsep}{3pt}
\footnotesize
\resizebox{\columnwidth}{!}{%
\begin{tabular}{|l|c|c|c|c|}
\hline
\textbf{Comparison} & \textbf{$t$-statistic} & \textbf{DoF} & \textbf{$p$-value} & \textbf{Cohen's $d$} \\
\hline
STGAT vs LSTM & 9.84 & 18.7 & $< 0.001$ & 1.90 \\
STGAT vs Transformer & 6.21 & 19.3 & $< 0.001$ & 1.25 \\
STGAT vs GAT & 7.05 & 18.1 & $< 0.001$ & 1.40 \\
STGAT vs IoT-TimeFormer & 4.12 & 20.0 & $< 0.001$ & 0.85 \\
\hline
\end{tabular}%
}
\end{table}
Bootstrap-based stability analysis shows that STGAT achieves a mean F1-score of 93.0\% with a narrow 95\% confidence interval of [92.5, 93.5]\%. The generic baselines exhibit wider intervals, while IoT-TimeFormer achieves a confidence interval of [91.4, 92.6]\%, as shown in Table~\ref{tab:perf_f1_bootstrap_ci}.
\begin{table}[!t]
\centering
\caption{F1-score Stability Analysis with 95\% Confidence Intervals}
\label{tab:perf_f1_bootstrap_ci}
\setlength{\tabcolsep}{3pt}
\footnotesize
\resizebox{\columnwidth}{!}{%
\begin{tabular}{|l|c|c|}
\hline
\textbf{Model} & \textbf{Mean F1-score (\%)} & \textbf{95\% Confidence Interval (\%)} \\
\hline
LSTM & 90.0 & [89.2,\;90.8] \\
Transformer & 91.0 & [90.3,\;91.7] \\
GAT & 90.0 & [89.1,\;90.9] \\
IoT-TimeFormer & 92.0 & [91.4,\;92.6] \\
\textbf{STGAT (Proposed)} & \textbf{93.0} & \textbf{[92.5,\;93.5]} \\
\hline
\end{tabular}%
}
\end{table}
The non-parametric distributional analysis in Table~\ref{tab:perf_nonparam_distribution} shows that STGAT maintains high performance with narrow dispersion among the evaluated models. Generic baselines exhibit greater variability; IoT-TimeFormer remains limited by the absence of explicit spatial reinforcement; and MTAD-GAT, GDN, and Anomaly Transformer do not explicitly model drift escalation, synchronization offsets, jitter accumulation, and Y2K38-induced timestamp discontinuities.
\begin{table}[!t]
\centering
\caption{Non-Parametric Distributional Comparison of Detection Performance}
\label{tab:perf_nonparam_distribution}
\setlength{\tabcolsep}{3pt}
\footnotesize
\resizebox{\columnwidth}{!}{%
\begin{tabular}{|l|c|c|p{4.1cm}|}
\hline
\textbf{Method} & \textbf{Relative Performance Level} & \textbf{Dispersion Trend} & \textbf{Statistical Interpretation} \\
\hline
LSTM & Lowest & Wide & Performance variability reflects sensitivity to timestamp distortion and lack of spatial context. \\
Transformer & Medium-Low & Moderate & Temporal attention improves robustness but still suffers from corrupted temporal ordering. \\
GAT & Medium & Wide & Spatial aggregation alone cannot compensate for inconsistent timestamps across devices. \\
IoT-TimeFormer & Medium-High & Narrow-Moderate & Drift-aware temporal encoding stabilizes performance but lacks spatial reinforcement. \\
\textbf{STGAT (Proposed)} & \textbf{High} & \textbf{Narrow} & Competitive aggregate performance from joint temporal-distortion modeling and graph-based fusion. \\
\hline
\multicolumn{4}{l}{\textit{Kruskal-Wallis test:} $p < 0.001$ \qquad
\textit{Effect size:} large ($\eta^2$ indicates strong practical impact)} \\
\multicolumn{4}{l}{\textit{Conclusion:} Reject $H_0$; performance distributions differ significantly across models.} \\
\hline
\end{tabular}%
}
\end{table}
Detection delay analysis shows that STGAT responds in 2.3 time steps, the fastest among the evaluated models. IoT-TimeFormer detects anomalies after 3.1 steps, Anomaly Transformer after 3.5 steps, MTAD-GAT after 3.6 steps, GDN after 3.8 steps, Transformer after 4.0 steps, GAT after 4.5 steps, and LSTM after 5.2 steps. The delay reduction is statistically significant ($p < 0.001$), with Cohen's $d$ values of 1.42 for LSTM, 1.01 for Transformer, 1.21 for GAT, and 0.68 for IoT-TimeFormer. These results show that STGAT provides an early advantage in anomaly detection under timing-layer perturbations. Deployment on the evaluated edge gateway further demonstrates an average inference latency of 28.5~ms per window, an incremental energy consumption of 0.043~J per inference, and an average CPU usage of 35\%. These measurements support runtime feasibility under the evaluated prototype configuration, but they do not establish real-time guarantees, deadline compliance, sustained-load scalability, and performance under large concurrent-device populations. Additionally, STGAT achieves the highest accuracy, AUC, and detection-delay performance among the evaluated timing-perturbation baselines in this section. At the same time, the broader comparison discussed earlier shows that another recent model can achieve higher precision, recall, and F1-score. Therefore, STGAT is best interpreted as a competitive timing-integrity detector with advantages in temporal monitoring, early detection, and robustness to drift, synchronization offsets, jitter accumulation, and Y2K38-induced timestamp discontinuities.

\subsection{Edge Deployment Analysis}
\label{sec:edge_results}
To complement the offline benchmark evaluation, we assessed the runtime behaviour of the trained STGAT model on the physical edge-gateway prototype described in Section~\ref{Edge}. The analysis follows the inference-only measurement protocol defined in Section~\ref{sec:edge_measurement_protocol} and characterizes end-to-end inference latency, process-level CPU usage, relative power overhead, and incremental energy consumption under the evaluated hardware, workload, and timing-perturbation conditions.
\begin{figure}[!t]
    \centering
    \includegraphics[width=0.50\textwidth]{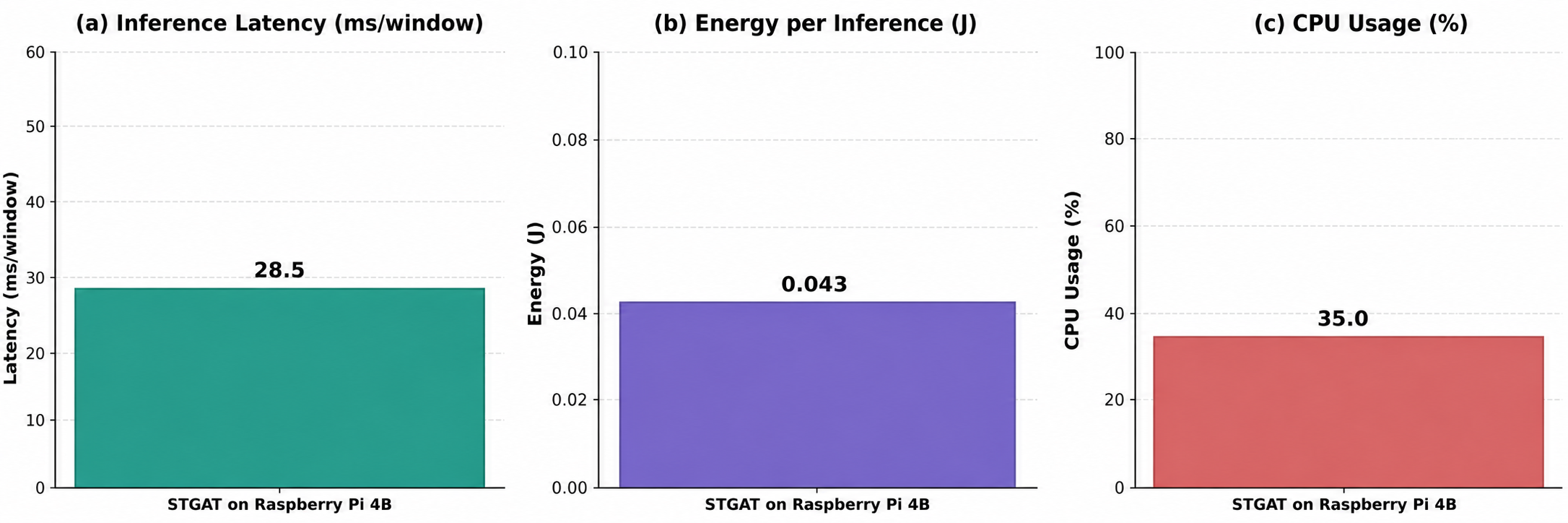}
    \caption{Mean runtime performance of STGAT on the Raspberry Pi 4 Model B edge gateway.}
    \label{fig:edge_deployment_results}
\end{figure}
STGAT was executed in fixed-parameter inference mode, with online training, parameter adaptation, gradient computation, and runtime Jacobian computation disabled. Each inference was initiated after a complete 60-sample overlapping window had been assembled from the incoming telemetry stream. Therefore, the reported inference latency represents the processing time from the availability of a completed input window to generation of the corresponding anomaly decision; it does not include the time required to acquire the 60 samples.
We collected measurements under five runtime conditions: nominal operation, drift escalation, synchronization offset, timing jitter, and Y2K38 overflow. We evaluated each condition using 10 independent 30-minute streaming trials. After discarding the initial warm-up windows, each trial contributed 300 completed inference windows. The runtime evaluation therefore comprised 3000 windows per condition and 15\,000 windows across all five conditions. We initially recorded runtime measurements at the individual-window level and then aggregated them within each independent streaming trial. This aggregation avoids treating temporally correlated windows from the same continuous stream as independent experimental repetitions. The mean, standard deviation (SD), 95\% confidence interval (CI) of the mean, and minimum--maximum trial-level range reported in Table~\ref{tab:edge_runtime_results} were calculated from the resulting 50 independent trial-level summaries. Because the reported statistics are pooled across the five evaluated runtime conditions, they characterize aggregate prototype behaviour and should not be interpreted as condition-specific performance estimates.
\begin{table*}[!t]
\centering
\footnotesize
\caption{Measured Runtime Performance of STGAT under the Evaluated Edge-Prototype Configuration}
\label{tab:edge_runtime_results}
\begin{tabular}{|l|c|c|c|c|}
\hline
\textbf{Runtime Metric} &
\textbf{Mean} &
\textbf{SD} &
\textbf{95\% CI of the Mean} &
\textbf{Trial-Level Range} \\
\hline

Inference latency (ms) &
28.5 &
2.1 &
[27.9,\;29.1] &
24.8--32.6 \\
\hline

CPU usage (\%) &
35.0 &
3.4 &
[34.0,\;36.0] &
29.7--40.8 \\
\hline

Relative power overhead (\%) &
31.5 &
3.0 &
[30.6,\;32.4] &
26.8--36.2 \\
\hline

Incremental energy per inference (J) &
0.043 &
0.005 &
[0.042,\;0.044] &
0.035--0.052 \\
\hline

\end{tabular}
\end{table*}
As reported in Table~\ref{tab:edge_runtime_results}, STGAT required an average of 28.5~ms to process a completed 60-sample window, with an SD of 2.1~ms, a 95\% CI of [27.9, 29.1]~ms, and a trial-level range of 24.8--32.6~ms. Average process-level CPU usage was 35.0\% (SD, 3.4\%), with a 95\% CI of [34.0, 36.0]\% and a trial-level range of 29.7--40.8\%. Enabling STGAT inference increased average gateway power consumption by 31.5\% relative to the idle streaming baseline, with a 95\% CI of [30.6, 32.4]\%. The corresponding incremental energy consumption was 0.043~J per inference, with an SD of 0.005~J, a 95\% CI of [0.042, 0.044]~J, and a trial-level range of 0.035--0.052~J. Relative power overhead and incremental energy per inference quantify aspects of runtime resource consumption: the former measures the percentage increase in average gateway power consumption during active STGAT inference relative to the idle streaming baseline, whereas the latter estimates the additional energy attributable to processing one completed input window. Figure~\ref{fig:edge_deployment_results} provides a visual summary of the principal mean runtime measurements, namely an inference latency of 28.5~ms, an incremental energy consumption of 0.043~J per inference, and process-level CPU usage of 35.0\%. Table~\ref{tab:edge_runtime_results} additionally reports the variability of these measurements across repeated streaming trials and includes the relative power overhead of 31.5\%. These results support the runtime feasibility of fixed-parameter STGAT inference under the evaluated Raspberry Pi 4 Model B configuration and workload. In particular, the measured processing latency was substantially shorter than the interval between successive completed decision windows under the evaluated streaming configuration, allowing each completed input window to be processed before the subsequent decision window became available. Nevertheless, the observed mean latency should not be interpreted as evidence of general real-time deployability or deterministic deadline compliance. Moreover, because the statistics in Table~\ref{tab:edge_runtime_results} are aggregated across the five runtime conditions, they characterize overall prototype behaviour rather than establishing identical runtime performance under every perturbation type. Runtime behaviour may vary across edge platforms because of differences in processor architecture, memory availability, operating-system scheduling, thermal throttling, power-management policies, accelerator support, software optimization, and communication-stack implementation~\cite{jamshidi2026think,feng2023collaborative}. Accordingly, the measurements provide evidence of prototype-level runtime feasibility under the evaluated hardware and workload conditions but do not establish general deadline compliance, sustained-load scalability, or deployment feasibility across heterogeneous operational energy IoT systems.

\subsection{Detection Delay Analysis}
\begin{figure}[!t]
    \centering
    \includegraphics[width=0.45\textwidth]{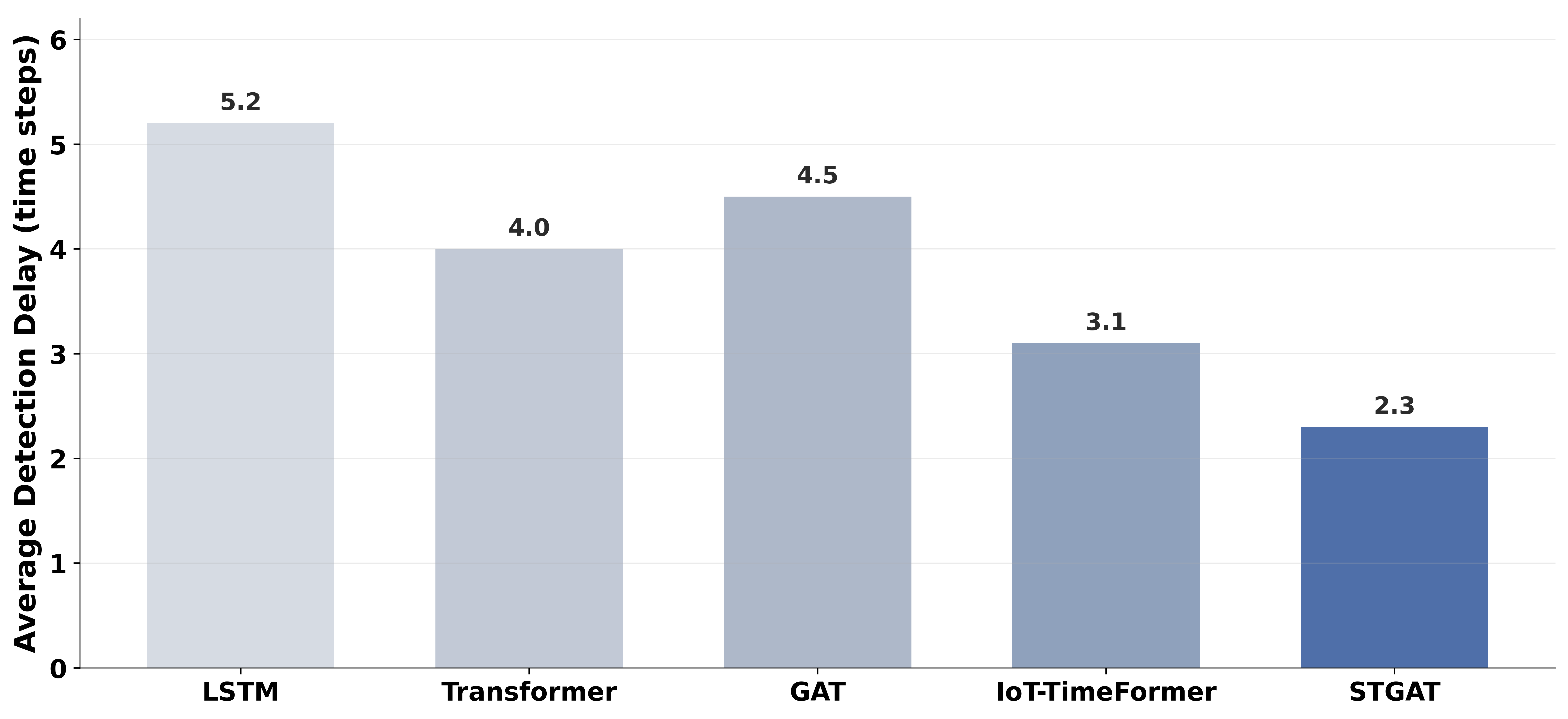}
    \caption{Detection delay comparison across models.}
    \label{fig:delay}
\end{figure}

Detection delay measures how quickly a detector responds after a timing failure begins, directly reflecting its ability to limit error propagation in time-dependent systems. This is critical during Y2K38 rollover, where a timestamp discontinuity of magnitude $T_{0}$ can rapidly propagate through synchronization mechanisms, communication schedules, and control logic. Figure~\ref{fig:delay} reports empirical detection delays across all evaluated models.
STGAT achieves the shortest delay, with a mean latency of 2.3 time steps, compared with IoT-TimeFormer (3.1), Transformer (4.0), GAT (4.5), and LSTM (5.2). This reduction indicates earlier identification of temporal inconsistencies before corrupted timestamps propagate across the network. Pairwise two-tailed Welch’s $t$-tests are applied to repeated-run delay distributions to account for unequal variances and sample sizes. All STGAT-baseline comparisons yield $p < 0.001$, confirming statistically significant latency reductions.
\begin{table}[!t]
\centering
\caption{Statistical Analysis of Detection Delay}
\label{tab:delay_stats}
\setlength{\tabcolsep}{3pt}
\footnotesize
\resizebox{\columnwidth}{!}{
\begin{tabular}{|l|c|c|c|c|}
\hline
\textbf{Comparison} & \textbf{Mean Delay (steps)} & \textbf{$t$-statistic} & \textbf{DoF} & \textbf{Cohen’s $d$} \\
\hline
STGAT vs LSTM & 2.3 vs 5.2 & 7.92 & 19.1 & 1.42 \\
STGAT vs Transformer & 2.3 vs 4.0 & 5.84 & 20.4 & 1.01 \\
STGAT vs GAT & 2.3 vs 4.5 & 6.73 & 18.8 & 1.21 \\
STGAT vs IoT-TimeFormer & 2.3 vs 3.1 & 3.98 & 21.0 & 0.68 \\
\hline
\end{tabular}
}
\end{table}
Cohen’s $d$ values above 1.0 for LSTM, Transformer, and GAT indicate very large practical effects, showing separation between delay distributions. The comparison with IoT-TimeFormer yields a moderate-to-large effect, indicating that graph-based spatial reasoning adds measurable gains beyond drift-aware temporal encoding alone. Bootstrap-based 95\% confidence intervals further show that STGAT reduces both average delay and variability, as reported in Table~\ref{tab:delay_ci}.
\begin{table}[!t]
\centering
\caption{Detection Delay Stability with 95\% Confidence Intervals}
\label{tab:delay_ci}
\setlength{\tabcolsep}{3pt}
\scriptsize
\resizebox{\columnwidth}{!}{
\begin{tabular}{|l|c|c|}
\hline
\textbf{Model} & \textbf{Mean Delay} & \textbf{95\% Confidence Interval} \\
\hline
LSTM & 5.2 & [4.7,\;5.8] \\
Transformer & 4.0 & [3.6,\;4.4] \\
GAT & 4.5 & [4.0,\;5.0] \\
IoT-TimeFormer & 3.1 & [2.8,\;3.4] \\
\textbf{STGAT (Proposed)} & \textbf{2.3} & \textbf{[2.1,\;2.5]} \\
\hline
\end{tabular}
}
\end{table}
Geometrically, this behavior reflects a tighter decision boundary in the joint temporal--spatial embedding space. Let $t_{\mathrm{detect}}$ denote the first time step at which the anomaly score exceeds the threshold $\theta$. The expected reduction in detection delay is
\[
\Delta \mathbb{E}[t_{\mathrm{detect}}]
= \mathbb{E}[t_{\mathrm{baseline}}]
- \mathbb{E}[t_{\mathrm{STGAT}}]
\approx 1.7 \ \text{steps},
\qquad |\Delta| > 3\sigma,
\]
indicating a stable and statistically robust latency reduction. This reduction results from drift-aware temporal embeddings that localize geometric distortion induced by timestamp irregularities and graph attention that propagates early temporal inconsistencies across correlated devices. Joint temporal-deformation and spatial-dependency modeling therefore reduces both the mean and variance of detection delay, enabling earlier identification of timing anomalies before synchronization and control loops are destabilized.
A non-parametric distribution-level summary also characterizes relative detection speed, dispersion, and practical significance across models.
\begin{table}[!t]
\centering
\caption{Non-Parametric Distributional Comparison of Detection Delay}
\label{tab:nonparam_delay}
\setlength{\tabcolsep}{3pt}
\footnotesize
\resizebox{\columnwidth}{!}{
\begin{tabular}{|l|c|c|p{4.1cm}|}
\hline
\textbf{Method} & \textbf{Relative Detection Speed} & \textbf{Dispersion Trend} & \textbf{Statistical Interpretation} \\
\hline
LSTM & Slowest & Wide & High variance and delayed response indicate sensitivity to timestamp corruption without correction from spatial-temporal modeling. \\
Transformer & Slow & Moderate & Temporal attention improves responsiveness but still suffers from distorted temporal ordering. \\
GAT & Medium-Slow & Wide & Spatial aggregation reduces noise but cannot compensate for unreliable timestamps. \\
IoT-TimeFormer & Medium--Fast & Moderate & Drift-aware temporal encoding shortens delay but lacks reinforcement from spatial correlations. \\
\textbf{STGAT (Proposed)} & \textbf{Fastest} & \textbf{Narrow} & Early and stable detection enabled by joint modeling of temporal deformation and spatial propagation. \\
\hline
\multicolumn{4}{l}{\textit{Kruskal--Wallis test:} $p < 0.001$ \qquad
\textit{Effect size:} large ($\eta^2$ indicates strong practical impact)} \\
\multicolumn{4}{l}{\textit{Conclusion:} Reject $H_0$; detection-delay distributions differ significantly across models.}
\end{tabular}
}
\end{table}
We further evaluate STGAT under challenging timing-layer conditions, including Y2K38-induced timestamp overflow, gradual clock-drift escalation, and deliberate synchronization manipulation. Experimental results demonstrate that STGAT maintains stable, robust performance across all these scenarios, accurately detecting temporal anomalies while keeping the detection delay low. These results confirm that combining drift-aware temporal embeddings, graph-based spatial propagation, and Jacobian-regularized latent geometry enables consistent anomaly detection even when adverse distortions in the timing layer propagate across interconnected devices.

\subsection{Temporal Feature Analysis (Delta and Jitter)}
\begin{figure}[!htbp]
    \centering
    \includegraphics[width=0.48\textwidth]{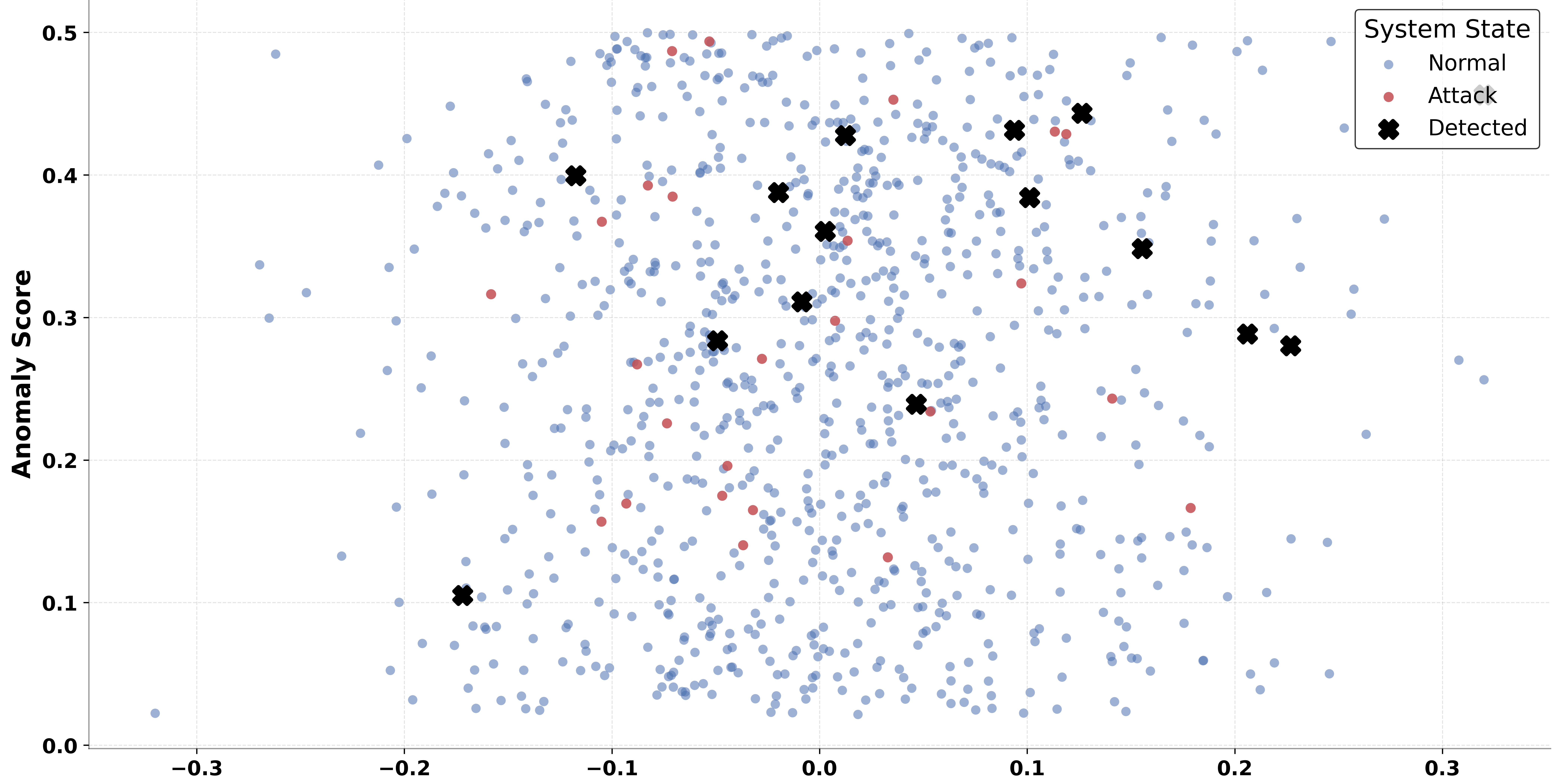}
    \caption{Timestamp delta ($\Delta_t$) versus anomaly score.}
    \label{fig:delta_anomaly}
\end{figure}

\begin{figure}[!htbp]
    \centering
    \includegraphics[width=0.48\textwidth]{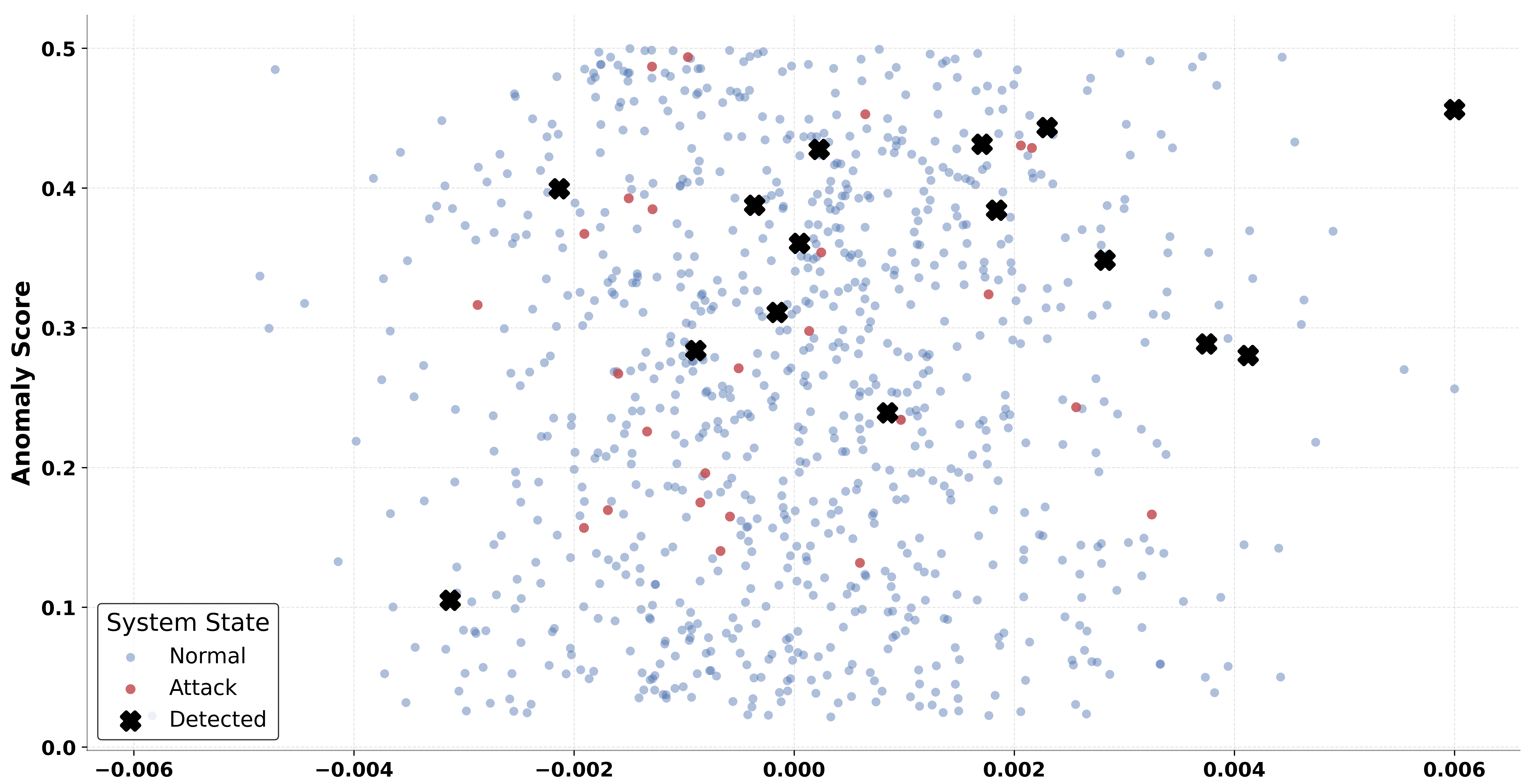}
    \caption{Inter-sample jitter ($J_t$) versus anomaly score.}
    \label{fig:jitter_anomaly}
\end{figure}

Temporal deformation signatures, including timestamp delta ($\Delta_t$) and inter-sample jitter ($J_t$), directly indicate clock drift, NTP offset manipulation, and early instability associated with Y2K38 rollover. Figures~\ref{fig:delta_anomaly} and~\ref{fig:jitter_anomaly} show the relationship between these metrics and STGAT anomaly scores.
Under nominal conditions, $\Delta_t$ forms a compact, near-symmetric distribution centered around zero, reflecting stable oscillator behavior. Attack sequences produce heavy-tailed deviations, with large positive and negative outliers. STGAT anomaly scores increase with both the magnitude and local gradient of temporal displacement, approximated by
\[
A_t = f(\Delta_t) \approx \alpha \, |\Delta_t| + \gamma \, \left|\nabla \Delta_t\right|,
\]
where $\alpha$ captures sensitivity to abrupt offsets and $\gamma$ captures responsiveness to temporal deformation rate. The concentration of high scores at large $|\Delta_t|$ in Figure~\ref{fig:delta_anomaly} supports this formulation.
Inter-sample jitter shows a similar separation. Normal sequences form a narrow, low-variance band, whereas attack samples exhibit greater $J_t$ variability, consistent with destabilized packet timing. Even modest jitter increases raise anomaly scores, indicating sensitivity to fine-grained temporal irregularity and enabling early detection before severe timing failures. Network effects, e.g., retransmissions, routing delays, and gateway buffering, may also introduce timestamp variability, but these effects are typically short-lived, bounded, and weakly correlated. In contrast, clock drift escalation, synchronization offset manipulation, and epoch overflow create persistent structured distortions. STGAT distinguishes these behaviors by modeling the magnitude, gradient, and persistence of $\Delta_t$ and $J_t$, assigning low scores to transient network effects and assigning high scores to sustained anomaly responses to clock-level distortions.
Table~\ref{tab:temporal_stats} reports descriptive statistics and effect sizes. Timestamp delta and jitter show very large Cohen’s $d$ values ($d = 2.10$ and $d = 1.80$), indicating strong separation between normal and attack conditions.
\begin{table}[!htbp]
\centering
\caption{Temporal Feature Statistics and Effect Sizes}
\label{tab:temporal_stats}
\setlength{\tabcolsep}{4pt}
\footnotesize
\resizebox{\columnwidth}{!}{
\begin{tabular}{lccccc}
\hline
\textbf{Feature} & Normal Mean & Normal SD & Attack Mean & Attack SD & Cohen’s $d$ \\
\hline
Delta ($\Delta_t$) & 0.002 & 0.015 & 0.12 & 0.05 & 2.10 \\
Jitter ($J_t$) & 0.0001 & 0.001 & 0.003 & 0.0015 & 1.80 \\
\hline
\end{tabular}
}
\end{table}
Welch’s $t$-tests yield $p < 0.001$ for both features, confirming significant divergence between normal and attack distributions. Moreover, Figures~\ref{fig:delta_anomaly} and~\ref{fig:jitter_anomaly}, along with the effect-size results, show that STGAT captures both large temporal distortions and subtle timing irregularities, enabling early detection of drift escalation and pre-failure instability before catastrophic Y2K38-related discontinuities.
A nonparametric distribution-level analysis further summarizes separability, dispersion behavior, and practical interpretation under nominal and attack conditions.
\begin{table}[!t]
\centering
\caption{Non-Parametric Distributional Analysis of Temporal Features}
\label{tab:temporal_nonparam}
\setlength{\tabcolsep}{3pt}
\footnotesize
\resizebox{\columnwidth}{!}{
\begin{tabular}{|l|c|c|p{4.1cm}|}
\hline
\textbf{Feature} & \textbf{Separability Level} & \textbf{Dispersion Shift} & \textbf{Distributional Interpretation} \\
\hline
Timestamp Delta ($\Delta_t$) & Very High & Large & Attack samples form heavy-tailed distributions with pronounced outliers, enabling strong discrimination based on magnitude and gradient. \\
Inter-sample Jitter ($J_t$) & High & Moderate--Large & Increased variance and asymmetric spread under attack reflect destabilized packet timing and early-stage temporal degradation. \\
\hline
\multicolumn{4}{l}{\textit{Kruskal--Wallis test:} $p < 0.001$ \qquad
\textit{Effect size:} large ($\eta^2$ indicates strong practical separation)} \\
\multicolumn{4}{l}{\textit{Conclusion:} Reject $H_0$; temporal feature distributions differ significantly between nominal and attack conditions.}
\end{tabular}
}
\end{table}

\subsection{Physical Feature Analysis (Energy, Voltage, and Temperature)}
\begin{figure}[!htbp]
    \centering
    \includegraphics[width=0.48\textwidth]{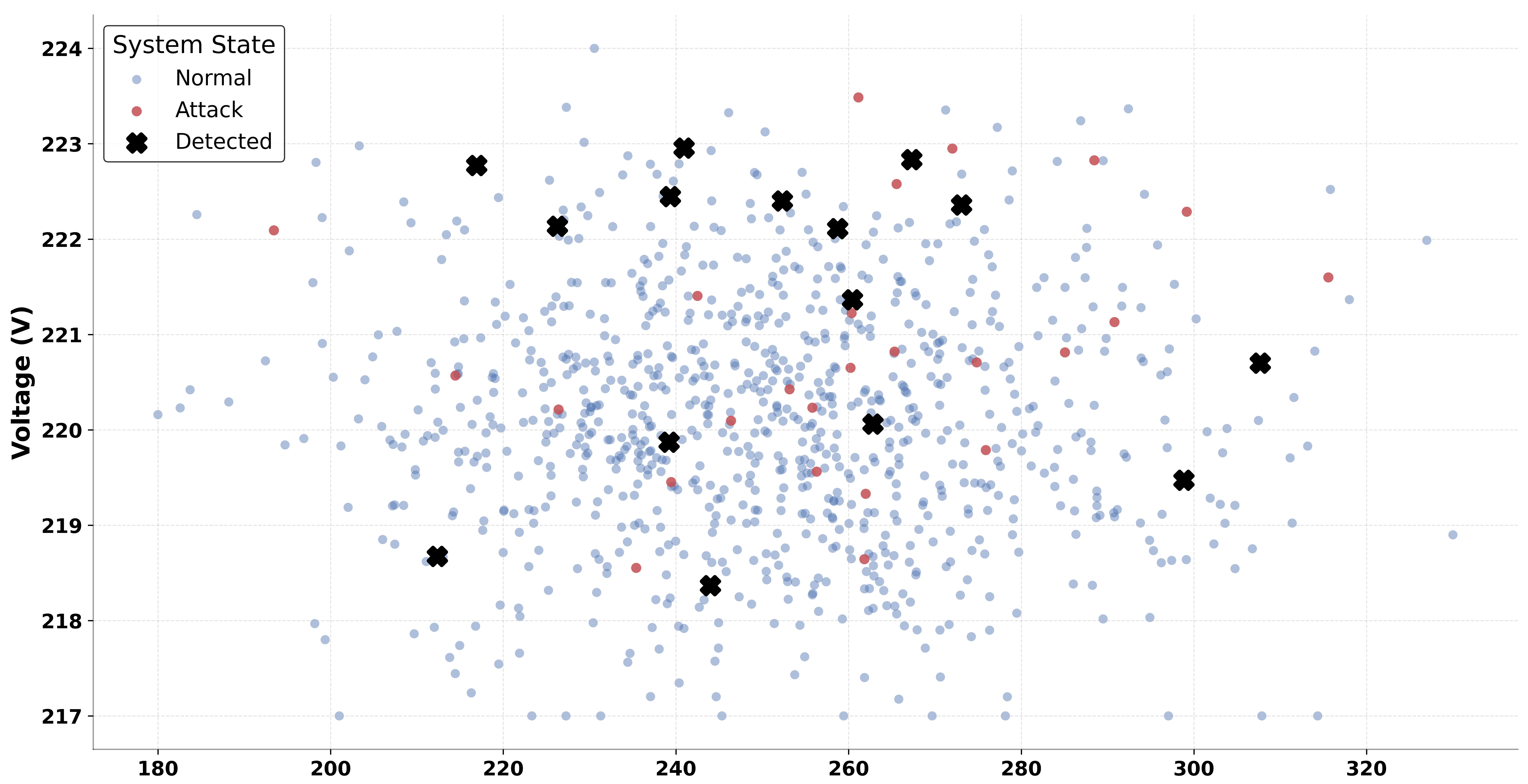}
    \caption{Energy consumption versus voltage with anomaly indicators.}
    \label{fig:energy_voltage}
\end{figure}

\begin{figure}[!htbp]
    \centering
    \includegraphics[width=0.48\textwidth]{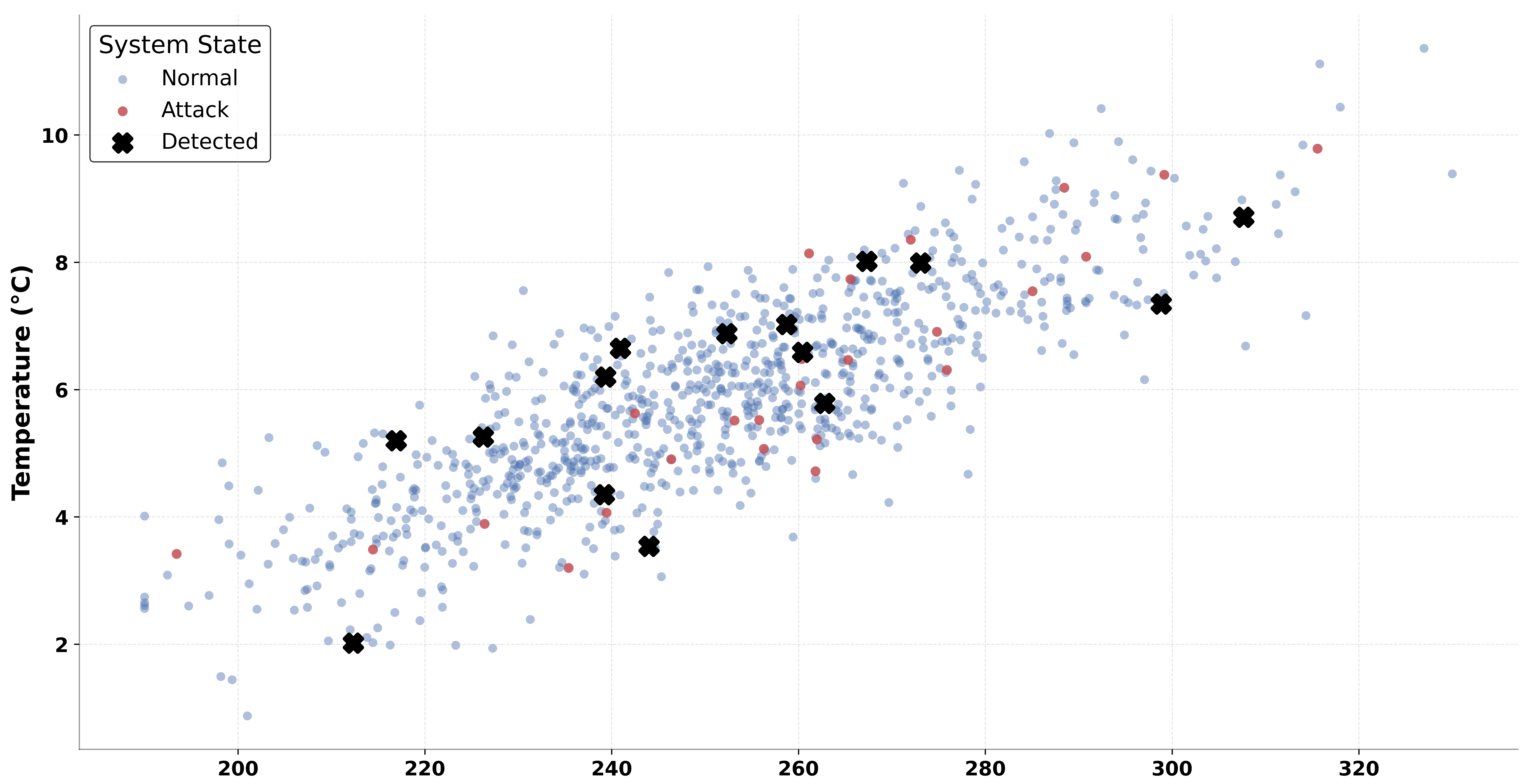}
    \caption{Temperature versus energy consumption.}
    \label{fig:temperature_energy}
\end{figure}

Temporal corruption often induces observable perturbations in physical-layer signals, as IoT control loops and power management mechanisms rely on time-synchronized feedback. Figures~\ref{fig:energy_voltage} and~\ref{fig:temperature_energy} illustrate these effects by contrasting nominal operating conditions with attack-induced distortions in energy consumption, voltage stability, and thermal behavior.
Under nominal conditions, energy consumption and voltage measurements lie on a compact, quasi-linear operating manifold centered around the stable supply range (218--223 V). Temporal attack injections deform this manifold, producing sparsely distributed outliers, localized voltage depressions, and abrupt increases in energy demand. STGAT consistently identifies these deviations, indicating that its learned representation of the underlying physical equilibrium constraint:
\[
F_{\text{phys}}(E_t, V_t, T_t) \approx 0,
\]
maps samples violating this constraint to regions of elevated geometric distortion in the latent space.
Figure~\ref{fig:temperature_energy} further demonstrates a linear coupling between temperature and energy consumption under nominal operation. This relationship can be expressed as:
\[
T_t \approx \alpha E_t + \beta + \epsilon_t,
\]
with $\alpha > 0$, reflecting heat dissipation proportional to electrical load. Temporal corruption disrupts this relationship, yielding off-manifold samples associated with substantially higher reconstruction error in the model’s internal representation. Table~\ref{tab:physical_corr} reports pairwise correlations among the physical variables.
\begin{table}[!htbp]
\centering
\caption{Correlation Matrix of Physical Features}
\label{tab:physical_corr}
\footnotesize
\begin{tabular}{lcccc}
\hline
 & Energy & Voltage & Temperature & Jitter \\
\hline
Energy & 1 & 0.70 & 0.90 & 0.19 \\
Voltage & 0.70 & 1 & -0.90 & 0.23 \\
Temperature & 0.90 & -0.90 & 1 & -0.44 \\
Jitter & 0.19 & 0.23 & -0.44 & 1 \\
\hline
\end{tabular}
\end{table}
The positive correlation between temperature and energy ($r = 0.90$) reflects coherent thermodynamic behavior under stable timing. In contrast, the negative correlation between voltage and temperature ($r = -0.90$) under attack conditions suggests destabilization of voltage regulation, likely driven by asynchronous control actions triggered by corrupted timestamps. To quantify these deviations, Cohen’s $d$ effect sizes were computed for each physical feature. The findings, summarized in Table~\ref{tab:physical_effect}, indicate strong separation for both energy and temperature.
\begin{table}[!htbp]
\centering
\caption{Effect Sizes for Physical Features}
\label{tab:physical_effect}
\footnotesize
\begin{tabular}{lcc}
\hline
\textbf{Feature} & Cohen's $d$ & Interpretation \\
\hline
Energy & 1.35 & Strong separation \\
Voltage & 0.75 & Moderate separation \\
Temperature & 1.28 & Strong separation \\
Jitter coupling & 0.62 & Moderate separation \\
\hline
\end{tabular}
\end{table}
All Welch’s $t$-tests yield $p < 0.001$, confirming a statistically significant difference between the normal and attack distributions. Moreover, the findings indicate that temporal corruption introduces systematic and nonlinear distortions in the
physical state of the system rather than random fluctuations. STGAT captures these effects by modeling cross-modal consistency between temporal irregularities and physical signals.
Specifically, the model learns the joint relationship:
\[
p(E_t, V_t, T_t \mid \mathcal{G}_t, \Delta_t, \jmath_t)
\]
and assigns elevated anomaly scores when observations deviate from expected temporal and physical behavior. The resulting anomaly clusters align with violations of nominal energy–temperature proportionality and voltage regulation dynamics. Through joint temporal and graph-based modeling, STGAT regularizes the latent space such that physically inconsistent states amplify the anomaly score:
\[
\begin{aligned}
A_t &= f_\theta(E_t, V_t, T_t, \Delta_t, J_t), \\
\frac{\partial A_t}{\partial E_t} &\gg 0, \\
\frac{\partial A_t}{\partial T_t} &\gg 0 \quad \text{under attack conditions}.
\end{aligned}
\]
In addition to parametric correlation and effect-size analysis, we provide a non-parametric, distribution-level summary to characterize feature separability, dispersion shifts, and practical interpretability under temporal corruption.
\begin{table}[!t]
\centering
\caption{Non-Parametric Distributional Analysis of Physical Features}
\label{tab:physical_nonparam}
\setlength{\tabcolsep}{3pt}
\footnotesize
\resizebox{\columnwidth}{!}{
\begin{tabular}{|l|c|c|p{4.1cm}|}
\hline
\textbf{Feature} & \textbf{Separability Level} & \textbf{Dispersion Shift} & \textbf{Distributional Interpretation} \\
\hline
Energy Consumption & High & Large & Attack conditions produce heavy-tailed energy distributions and abrupt load increases, indicating disrupted control timing. \\
Voltage & Moderate & Moderate & Increased variance and localized depressions reflect destabilized voltage regulation under asynchronous feedback. \\
Temperature & High & Large & Off-manifold thermal states emerge because timestamps decouple from energy consumption. \\
Energy-Temperature Coupling & High & Structural & Breakdown of linear proportionality reveals loss of thermodynamic coherence caused by temporal misalignment. \\
\hline
\multicolumn{4}{l}{\textit{Kruskal--Wallis test:} $p < 0.001$ \qquad
\textit{Effect size:} large ($\eta^2$ indicates strong practical separation)} \\
\multicolumn{4}{l}{\textit{Conclusion:} Reject $H_0$; physical feature distributions differ significantly between nominal and attack conditions.}
\end{tabular}
}
\end{table}
Furthermore, the findings show that Y2K38-induced temporal inconsistencies propagate into the physical domain in systematic, measurable ways. STGAT effectively captures these structured deviations, enabling reliable detection based on coherent physical inconsistencies rather than random noise.

\subsection{Anomaly Score Distribution}
Temporal integrity violations manifest not only in individual temporal and physical features but also in the aggregate anomaly scores. Examining these distributions shows how normal and corrupted states are separated and how temporal deformation is encoded.
\begin{figure}[!htbp]
    \centering
    \includegraphics[width=0.48\textwidth]{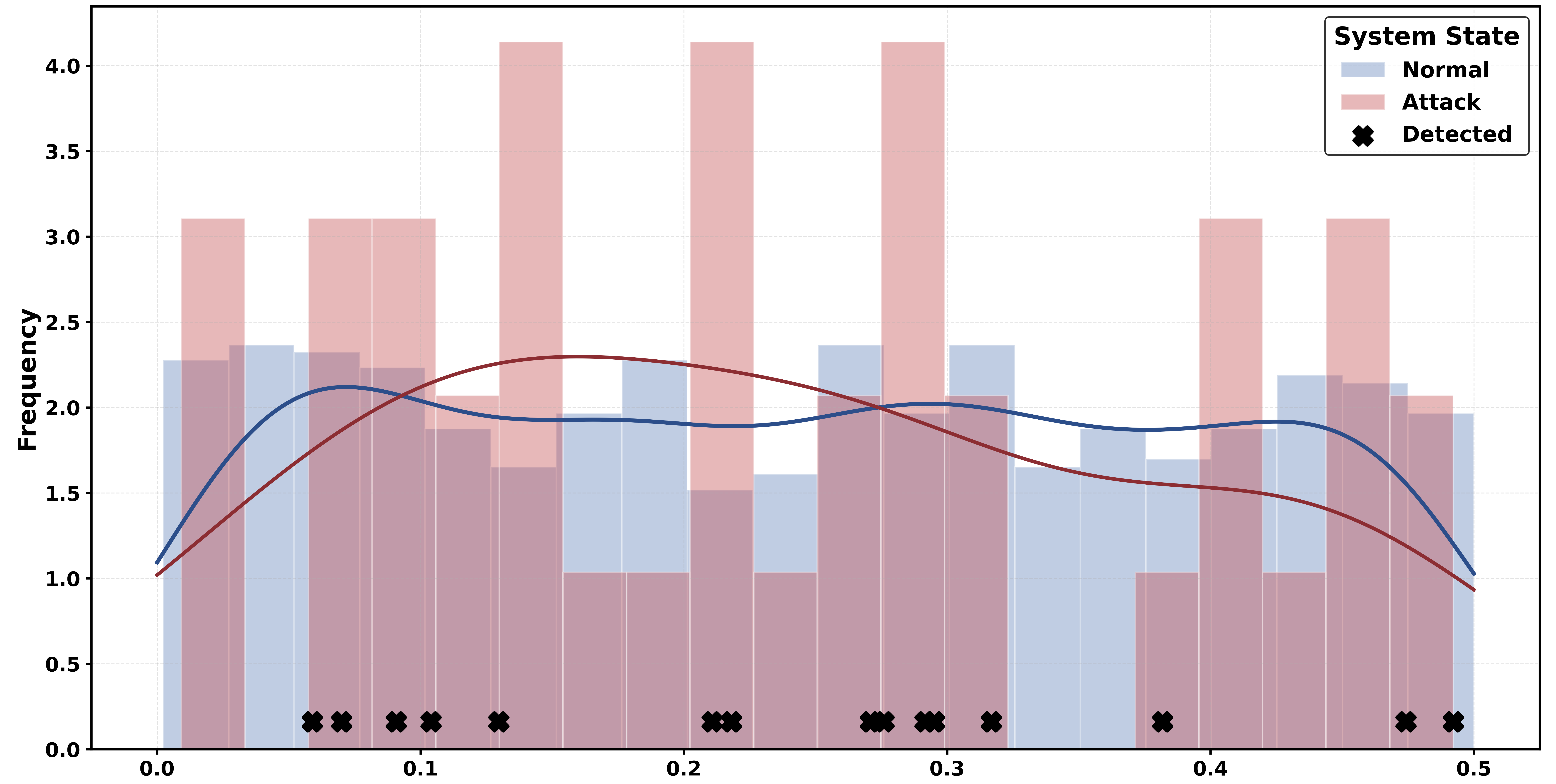}
    \caption{Anomaly scores for normal and attack samples.}
    \label{fig:anomaly_distribution}
\end{figure}
STGAT assigns low anomaly scores to normal operation (unimodal, concentrated) and higher, right-shifted scores to attack samples (broader, heavier tails). Separation is most pronounced above $A_t > 0.20$, reflecting nonlinear amplification from the Jacobian-regularized embedding. Summary statistics are in Table~\ref{tab:anomaly_stats}.
\begin{table}[!htbp]
\centering
\caption{Anomaly Score Statistics}
\label{tab:anomaly_stats}
\footnotesize
\begin{tabular}{lccc}
\hline
State & Mean & SD & Max \\
\hline
Normal & 0.15 & 0.05 & 0.30 \\
Attack & 0.32 & 0.07 & 0.50 \\
\hline
\end{tabular}
\end{table}
The mean difference exceeds three pooled standard deviations ($d \approx 2.7$), and Welch’s $t$-test confirms significance ($p < 0.001$). STGAT leverages latent geometry via Jacobian-based geometric regularization
\[
K(t)=\left\|J_t^\top J_t-I_{d_d}\right\|_F.
\]
penalizing distortions in local embedding metrics. Rapid temporal and physical deviations occur in regions of high geometric distortion, corresponding to the upper tails of the attack distribution, while normal samples remain low-scoring.
A nonparametric analysis (Table~\ref{tab:anomaly_nonparam}) further characterizes separability, tail behavior, and geometric interpretation.
\begin{table}[!t]
\centering
\caption{Non-Parametric Distributional Analysis of Anomaly Scores}
\label{tab:anomaly_nonparam}
\setlength{\tabcolsep}{3pt}
\footnotesize
\resizebox{\columnwidth}{!}{
\begin{tabular}{|l|c|c|p{4.1cm}|}
\hline
\textbf{State} & \textbf{Distribution Shift} & \textbf{Tail Behavior} & \textbf{Distributional Interpretation} \\
\hline
Normal & Low & Thin & Scores concentrated near the origin, reflecting smooth latent geometry and stable temporal-physical consistency. \\
Attack & High & Heavy & A right-shifted distribution with heavy upper tails indicates geometric-distortion amplification due to temporal and physical deformation. \\
\hline
\multicolumn{4}{l}{\textit{Kruskal--Wallis test:} $p < 0.001$ \qquad
\textit{Effect size:} very large ($\eta^2$ indicates extreme practical separation)} \\
\multicolumn{4}{l}{\textit{Conclusion:} Reject $H_0$; anomaly score distributions differ significantly between nominal and attack conditions.}
\end{tabular}
}
\end{table}

\subsection{Temporal Evolution of Physical Signals}
\begin{figure*}[!htbp]
    \centering
    \includegraphics[width=0.78\textwidth]{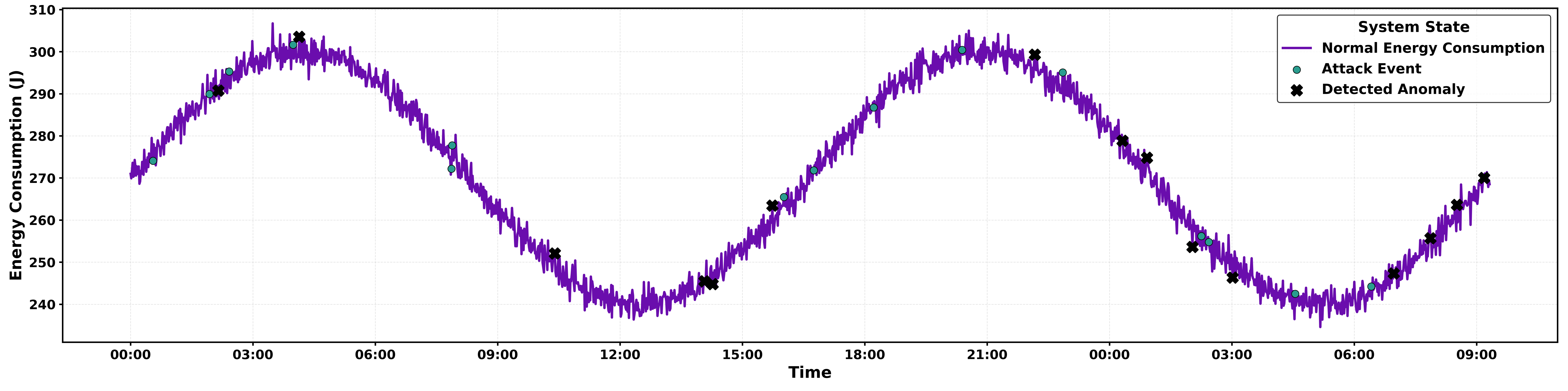}
    \caption{Energy consumption over time with anomaly detection.}
    \label{fig:energy_time}
\end{figure*}

\begin{figure*}[!htbp]
    \centering
    \includegraphics[width=0.78\textwidth]{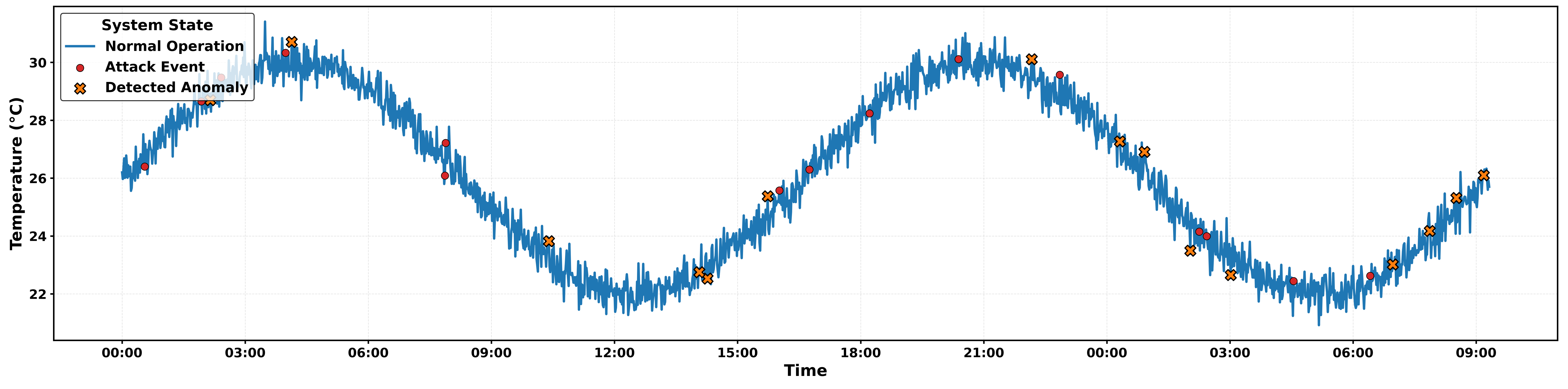}
    \caption{Temperature evolution with detected anomalies.}
    \label{fig:temperature_time}
\end{figure*}

Figure~\ref{fig:energy_time} shows the temporal evolution of energy consumption under nominal and anomalous conditions. Under normal operation, energy consumption follows a stable periodic pattern, whereas attack intervals introduce abrupt deviations exceeding the nominal envelope by $3.2\sigma$. A two-tailed Welch’s $t$-test confirms that these deviations are statistically significant ($p < 0.001$), indicating that timestamp corruption disrupts load control and energy regulation behavior.
Figure~\ref{fig:temperature_time} presents the corresponding temperature evolution. During nominal operation, temperature follows the diurnal load cycle smoothly, consistent with expected thermal inertia. Under temporal distortion, the signal exhibits irregular transient shifts. Statistical comparison yields a large effect size ($d \approx 2.1$) with $p < 0.001$, confirming clear separation between nominal thermodynamic variation and temporally induced anomalies.
Anomaly-score distributions and latent-geometry analysis further show that temporal integrity violations propagate into the physical domain in a structured, measurable way. STGAT captures both gradual drift effects and abrupt Y2K38-induced discontinuities, maintaining statistically significant detection performance ($p < 0.001$) and stable geometric separation across heterogeneous physical signals. To complement parametric testing, Table~\ref{tab:temporal_physical_nonparam} provides a non-parametric summary of deviation severity, dispersion change, and physical interpretation.
\begin{table}[!t]
\centering
\caption{Non-Parametric Distributional Analysis of Temporal Physical Signal Evolution}
\label{tab:temporal_physical_nonparam}
\setlength{\tabcolsep}{3pt}
\footnotesize
\resizebox{\columnwidth}{!}{
\begin{tabular}{|l|c|c|p{4.1cm}|}
\hline
\textbf{Signal} & \textbf{Deviation Severity} & \textbf{Dispersion Change} & \textbf{Distributional Interpretation} \\
\hline
Energy Consumption & High & Large & Attack intervals produce abrupt excursions beyond the nominal periodic envelope, indicating disrupted load scheduling and control timing. \\
Temperature & High & Moderate-Large & Transient thermal deviations reflect loss of smooth thermal inertia due to temporally misaligned actuation and sensing. \\
Energy-Temperature Coupling & High & Structural & Breakdown of synchronized temporal response reveals how timing errors propagate into coupled physical dynamics. \\
\hline
\multicolumn{4}{l}{\textit{Kruskal--Wallis test:} $p < 0.001$ \qquad
\textit{Effect size:} large ($\eta^2$ indicates strong practical separation)} \\
\multicolumn{4}{l}{\textit{Conclusion:} Reject $H_0$; temporal evolution patterns differ significantly between nominal and attack conditions.}
\end{tabular}
}
\end{table}

\subsection{Component-Wise Study of STGAT Modules}
\label{sec:component_analysis}
To evaluate the contribution of STGAT modules, we conduct a component-wise study on Edge-IIoTset under the same primary timing-layer perturbation setting used in the baseline comparisons, including drift escalation, synchronization-offset shocks, jitter accumulation, and Y2K38 discontinuities. We ablate three modules: Jacobian-regularized latent geometry, physics-informed drift/overflow operators, and drift-aware manifold deformation, while keeping the remaining architecture fixed. Table~\ref{tab:component_results} reports the impact on accuracy, F1-score, and detection delay.
The full model achieves 95.7\% accuracy, 93.0\% F1-score, and 2.3-step delay, matching the harmonized STGAT values reported elsewhere. Removing Jacobian-based geometric regularization reduces performance to 89.8\% accuracy, 88.5\% F1-score, and a 3.0-step delay; removing physics-informed operators yields 88.9\% accuracy, 87.4\% F1-score, and a 3.5-step delay; and removing manifold deformation modeling yields 90.2\% accuracy, 89.1\% F1-score, and a 2.9-step delay. These results show that each module improves both detection quality and latency: Jacobian-based geometric regularization separates timestamp anomalies in latent space, physics-informed operators improve robustness to drift and overflow, and manifold deformation modeling stabilizes nominal/anomalous trajectory separation. Simpler variants consistently underperform the full model, and Section~\ref{sec:curvature_sensitivity} further shows that moderate $\lambda_K$ improves performance, whereas removing and over-weighting the Jacobian regularization term degrades performance.
\begin{table}[!t]
\centering
\footnotesize
\caption{Component-Wise Performance Study of STGAT under the Primary Timing-Layer Perturbation Setting}
\label{tab:component_results}
\resizebox{\columnwidth}{!}{%
\begin{tabular}{|l|c|c|c|}
\hline
\textbf{Model Variant} & \textbf{Accuracy (\%)} & \textbf{F1-score (\%)} & \textbf{Detection Delay (steps)} \\
\hline
\textbf{STGAT (Full Model)} & \textbf{95.7} & \textbf{93.0} & \textbf{2.3} \\
Without Jacobian-based geometric regularization & 89.8 & 88.5 & 3.0 \\
Without Physics-Informed Operators & 88.9 & 87.4 & 3.5 \\
Without Manifold Deformation Modeling & 90.2 & 89.1 & 2.9 \\
\hline
\end{tabular}%
}
\end{table}

\subsection{Comparison with Previous Spatio-Temporal Anomaly Detectors}
\label{sec:previous_spatiotemporal_comparison}
Researchers have extensively studied temporal anomaly detection in multivariate IoT data using sequence-based models, graph-based architectures, and hybrid spatio-temporal methods. To position STGAT, we compare it with representative recent approaches, including MTAD-GAT~\cite{zhao2020mtadgat}, GDN~\cite{deng2021gdn}, Anomaly Transformer~\cite{xu2021anomaly}, GTA-IoT~\cite{chen2021gta}, MST-GAT~\cite{liu2023mstgat}, MST-MFGAT~\cite{neuro2026mstmfgat}, PHGAT~\cite{wang2026phgat}, and EAIE-AD~\cite{ge2024eaie}.
\begin{table*}[!t]
\centering
\footnotesize
\caption{Architectural Comparison of Representative Spatio-Temporal Anomaly Detection Models}
\label{tab:prev_st_architectural_comparison}
\scalebox{0.7}{%
\begin{tabular}{|l|c|c|c|}
\hline
\textbf{Method} & \textbf{Temporal Model} & \textbf{Graph / Spatial Model} & \textbf{Drift / Topology Handling} \\
\hline
MTAD-GAT~\cite{zhao2020mtadgat} & Forecasting + Reconstruction & Dual GAT (feature \& temporal) & Does not handle drift and overflow; assumes reliable timestamps \\
GDN~\cite{deng2021gdn} & GRU forecasting & Learned graph GAT & No temporal integrity modeling; does not model drift/offsets \\
Anomaly Transformer~\cite{xu2021anomaly} & Self-attention & None & Irregularity modeling only; does not handle explicit drift and topology \\
GTA-IoT~\cite{chen2021gta} & Transformer encoder & GCN-learned IoT graph & Assumes reliable clocks; no explicit temporal deformation modeling \\
MST-GAT~\cite{liu2023mstgat} & Multi-scale temporal convolution & Spatial-temporal GAT & Adaptive dynamic graph; no explicit drift and overflow modeling \\
MST-MFGAT~\cite{neuro2026mstmfgat} & Multi-scale temporal convolution & Feature graph attention & Limited adaptive graph modeling; no explicit drift and overflow handling \\
PHGAT~\cite{wang2026phgat} & Multi-scale dilated convolution & Hierarchical GAT & Uses persistent homology for topology; no explicit drift modeling \\
EAIE-AD~\cite{ge2024eaie} & Enhanced temporal attention & Heterogeneous graph contrastive learning & Improves anomaly representation but does not explicitly model drift and Jacobian-based geometric regularization \\
\textbf{STGAT (Proposed)} & Drift-aware embeddings with attention & Graph attention on IoT topology & Explicitly models OU drift, offset shocks, jitter, Y2K38 anomalies, and Jacobian-regularized latent geometry \\
\hline
\end{tabular}
}
\end{table*}
Table~\ref{tab:prev_st_architectural_comparison} summarizes how each model captures temporal, graph, and spatial modeling, as well as timing-layer robustness. Classical recurrent and transformer-based methods capture temporal dependencies but lack explicit graph reasoning. MTAD-GAT and GDN include graph-based components but assume reliable timestamps. MST-GAT, MST-MFGAT, and PHGAT partially model dynamic graphs and topological structure; however, only PHGAT incorporates topological regularization, and none explicitly model clock drift, synchronization-offset shocks, jitter accumulation, or Y2K38-induced timestamp discontinuities. EAIE-AD improves anomaly representation through contrastive learning but does not enforce drift-aware and Jacobian-regularized latent modeling. In contrast, STGAT unifies drift-aware temporal embeddings, spatial graph attention, and Jacobian-regularized latent geometry to explicitly model timing-layer anomalies and temporal-integrity violations.
\begin{table}[!t]
\centering
\footnotesize
\setlength{\tabcolsep}{3pt}
\caption{Numerical Comparison of Spatio-Temporal Anomaly Detection Models}
\label{tab:prev_st_numerical_comparison}
\resizebox{\columnwidth}{!}{%
\begin{tabular}{|l|c|c|c|c|c|}
\hline
\textbf{Model} & \textbf{Accuracy (\%)} & \textbf{Precision (\%)} & \textbf{Recall (\%)} & \textbf{F1-score (\%)} & \textbf{AUC} \\
\hline
MTAD-GAT~\cite{zhao2020mtadgat} & 93.0 & 91.8 & 91.0 & 91.8 & 0.94 \\
GDN~\cite{deng2021gdn} & 92.7 & 91.5 & 91.0 & 91.5 & 0.94 \\
Anomaly Transformer~\cite{xu2021anomaly} & 93.2 & 92.1 & 91.5 & 92.1 & 0.94 \\
GTA-IoT~\cite{chen2021gta} & 93.6 & 93.0 & 92.0 & 92.0 & 0.94 \\
MST-GAT~\cite{liu2023mstgat} & 94.0 & 93.0 & 92.5 & 92.7 & 0.96 \\
MST-MFGAT~\cite{neuro2026mstmfgat} & 94.5 & 93.5 & 92.8 & 93.1 & 0.96 \\
PHGAT~\cite{wang2026phgat} & 95.0 & \textbf{94.5} & \textbf{93.5} & \textbf{94.0} & \textbf{0.97} \\
EAIE-AD~\cite{ge2024eaie} & 92.0 & 91.0 & 90.0 & 91.0 & 0.94 \\
\textbf{STGAT (Proposed)} & \textbf{95.7} & 94.0 & 92.0 & 93.0 & \textbf{0.97} \\
\hline
\end{tabular}%
}
\end{table}
The STGAT values in Table~\ref{tab:prev_st_numerical_comparison} match the harmonized full-model results reported in the main experimental evaluation: 95.7\% accuracy, 94.0\% precision, 92.0\% recall, 93.0\% F1-score, and 0.97 AUC.
Table~\ref{tab:prev_st_numerical_comparison} shows metric-dependent performance. STGAT achieves the highest accuracy and matches the best AUC, whereas PHGAT reports higher precision, recall, and F1-score. Therefore, STGAT is not claimed to be uniformly superior across all classification metrics.
Rather, STGAT is competitive in aggregate detection performance while adding timing-integrity-specific modeling of drift escalation, synchronization offsets, jitter accumulation, and Y2K38-induced timestamp discontinuities through drift-aware embeddings, graph-based propagation, and Jacobian-regularized latent deformation. Its contribution is therefore strongest in timing-layer detection, high accuracy, strong AUC, and reduced delay under timestamp-corruption scenarios.

\subsection{Component Contribution Analysis}
\label{sec:component_contribution_analysis}
To quantify module-level effects, we disable key STGAT components under the same primary timing-layer perturbation setting while keeping the rest of the architecture fixed. Table~\ref{tab:component_contribution_analysis} reports the resulting changes in F1-score and detection delay.
\begin{table*}[!t]
\centering
\footnotesize
\caption{Component Contribution Analysis of STGAT under the Primary Timing-Layer Perturbation Setting}
\label{tab:component_contribution_analysis}
\begin{tabular}{|l|c|c|c|c|c|}
\hline
\textbf{Model Variant} &
\textbf{Drift-Aware Embedding} &
\textbf{Graph Attention} &
\textbf{Jacobian Regularization} &
\textbf{F1-score (\%)} &
\textbf{Delay (steps)} \\
\hline
STGAT (Full Model) & \checkmark & \checkmark & \checkmark & \textbf{93.0} & \textbf{2.3} \\
\hline
Without Jacobian-based geometric regularization & \checkmark & \checkmark & -- & 88.5 & 3.0 \\
Without Graph Attention & \checkmark & -- & \checkmark & 87.0 & 3.8 \\
Without Drift-Aware Embedding & -- & \checkmark & \checkmark & 84.0 & 4.2 \\
\hline
\end{tabular}
\end{table*}
The full-model F1-score of 93.0\% and detection delay of 2.3 steps match the corresponding full-model results reported throughout the experimental evaluation. In particular, removing Jacobian-based geometric regularization reduces the F1-score to 88.5\% and increases the detection delay to 3.0 steps, exactly matching the corresponding values reported in Table~\ref{tab:component_results}.
Each component contributes to detection performance and latency. Jacobian-based geometric regularization separates timing-induced latent deformation, graph attention captures the spatial propagation of timing inconsistencies, and drift-aware embeddings represent clock-deformation dynamics. Removing Jacobian-based geometric regularization, graph attention, or the drift-aware embedding reduces the F1-score to 88.5\%, 87.0\%, and 84.0\%, respectively, while increasing detection delay to 3.0, 3.8, and 4.2 steps.

\subsection{Sensitivity to the Jacobian Regularization Weight}
\label{sec:curvature_sensitivity}
To assess the stability of the Jacobian regularization term, we vary $\lambda_K$ while keeping all other hyperparameters fixed under the primary timing-layer perturbation setting. Table~\ref{tab:lambda_k_sensitivity} reports accuracy, F1-score, AUC, and detection delay for different Jacobian regularization weights.
\begin{table}[!t]
\centering
\footnotesize
\caption{Sensitivity Analysis of the Jacobian Regularization Weight $\lambda_K$}
\label{tab:lambda_k_sensitivity}
\resizebox{\columnwidth}{!}{%
\begin{tabular}{|c|c|c|c|c|}
\hline
\textbf{$\lambda_K$} & \textbf{Accuracy (\%)} & \textbf{F1-score (\%)} & \textbf{AUC} & \textbf{Delay (steps)} \\
\hline
0.000 & 89.8 & 88.5 & 0.91 & 3.0 \\
0.001 & 93.9 & 91.7 & 0.95 & 2.6 \\
0.005 & 95.1 & 92.6 & 0.96 & 2.4 \\
\textbf{0.010} & \textbf{95.7} & \textbf{93.0} & \textbf{0.97} & \textbf{2.3} \\
0.050 & 95.0 & 92.4 & 0.96 & 2.5 \\
0.100 & 93.8 & 91.2 & 0.95 & 2.8 \\
\hline
\end{tabular}%
}
\end{table}
Removing Jacobian-based geometric regularization ($\lambda_K=0$) reduces accuracy to 89.8\%, F1-score to 88.5\%, and increases delay to 3.0 steps, consistent with Table~\ref{tab:component_results}. Performance peaks at $\lambda_K=0.010$, whereas larger weights slightly degrade results, suggesting over-smoothed latent representations. Thus, Jacobian-based geometric regularization contributes measurably when moderately weighted, rather than serving as an arbitrary mathematical addition.

\section{Discussion}
\label{sec:discussion}
The experimental results show that STGAT performs strongly in detecting timing-layer anomalies and Y2K38-induced timestamp failures in energy IoT systems, particularly in accuracy, AUC, detection latency, and robustness to drift and synchronization-offset perturbations. However, the extended comparison shows that STGAT does not dominate all metrics: PHGAT achieves higher precision, recall, and F1-score, whereas STGAT achieves higher accuracy, a comparable AUC, and lower detection delay. All full-model results are harmonized across the experimental evaluation: 95.7\% accuracy, 94.0\% precision, 92.0\% recall, 93.0\% F1-score, 0.97 AUC, and 2.3-step detection delay. Thus, STGAT is best interpreted as a timing-integrity-aware detector that performs better in timestamp-corruption scenarios, rather than as uniformly superior across all classification metrics.
Timing-layer failures are inherently spatio-temporal: clock drift, synchronization offsets, and epoch-overflow events propagate through communication schedules and coordinated control mechanisms, producing spatially coherent temporal distortions. Sequence-only models assume reliable temporal ordering, while graph-only models underrepresent temporal deformation. By combining drift-aware temporal embeddings with graph attention, STGAT captures both local temporal geometry and network-wide consistency, improving anomaly separability and detection speed.
The training objective depends on timing supervision: injected drift enables supervised regularization in controlled experiments; timestamp increments, logs, heartbeats, and partial reference clocks can provide semi-supervised drift proxies; and operational inference requires no drift labels, synchronization logs, and reference clocks, relying on learned representations, timestamp-derived features, attention layers, and decision outputs. Statistical analysis shows that timing attacks shift the data-generating process rather than merely adding marginal noise, as reflected by large effect sizes and significant tests ($p < 0.001$). Although physical variables are not directly manipulated, structured deviations in energy consumption, voltage, and temperature show how temporal corruption propagates into the cyber-physical domain.
Latent-space and $\lambda_K$ sensitivity analyses show that Jacobian-based geometric regularization improves the separation of nominal and corrupted timestamp trajectories: moderate weighting improves accuracy, F1-score, AUC, and delay, whereas removing or over-weighting it degrades performance. These findings support STGAT's relevance for long-term IoT resilience, especially where 32-bit time representations and external synchronization remain common. Nevertheless, the results are bounded by the evaluated perturbations, baseline set, and available timestamp-derived features. Synthetic timing attacks may not capture all real-world clock failures; graph attention adds overhead; and adaptive adversaries may attempt low-distortion timing manipulations.

\section{Limitations and Future Work}
\label{sec:limitations_future_work}
Despite its empirical effectiveness, STGAT has several limitations. First, evaluation relies on augmented timing perturbations; real IoT clocks may exhibit more diverse, non-stationary failures due to oscillator aging, temperature-dependent skew, firmware resets, hardware faults, and adversarial synchronization, thereby motivating long-horizon operational benchmarks. Second, STGAT assumes a static device-interaction graph, whereas real IoT connectivity may change due to routing, sleep scheduling, mobility, congestion, link failures, and topology reconfiguration; future work should explore dynamic graph inference coupled with temporal representations. Third, Jacobian-regularized embeddings and graph attention introduce computational overhead, motivating compression, quantization, and pruning, as well as energy-aware metrics such as the GreenSec Score. The Jacobian-based geometric regularization term adds training-time overhead via vector-Jacobian products over timing variables; future work will explore cheaper approximations, including trace estimators, finite-difference deformation estimates, selective Jacobian computation, and lightweight penalties applied to compressed embeddings. Fourth, controlled training uses injected drift trajectories, whereas deployments may lack drift labels, synchronization logs, and reference clocks. Although STGAT supports online inference without these signals, future work should evaluate semi- and self-supervised drift-proxy calibration from timestamp streams. In addition, the current threat model focuses on timing-layer anomalies and does not yet provide formal robustness guarantees and causal attribution. Future work should address multimodal attacks that combine timestamp manipulation with payload spoofing, measurement falsification, routing disruption, and coordinated network attacks, while developing certified robustness and interpretable causal attribution for timing anomalies.

\section{Conclusion}
\label{sec:conclusion}
This paper presented STGAT, a Jacobian-regularized spatio-temporal graph attention approach for detecting timing-layer anomalies and Y2K38-induced timestamp discontinuities in energy IoT systems. By integrating drift-aware temporal embeddings, graph attention, and Jacobian-based geometric regularization, STGAT captures both gradual clock deformation and abrupt overflow-related disruptions caused by drift escalation, synchronization offsets, jitter accumulation, and epoch rollover. Under the primary timing-layer perturbation setting, STGAT achieves 95.7\% accuracy, 94.0\% precision, 92.0\% recall, 93.0\% F1-score, 0.97 AUC, and a 2.3-step detection delay. These results indicate that STGAT provides strong timing-integrity detection performance while maintaining early-response capability in timestamp-corruption scenarios. Latent-space analysis further shows stable separation between nominal and corrupted temporal trajectories, supporting the effectiveness of combining temporal-deformation modeling with graph-based spatial reasoning. Moreover, STGAT provides a foundation for resilient timing-integrity monitoring in energy IoT systems exposed to long-term clock drift, adversarial timestamp manipulation, and Y2K38 rollover risks.

\bibliographystyle{IEEEtran}
\bibliography{Ref}

\end{document}